\documentclass[pdflatex,sn-mathphys-ay]{sn-jnl}

\usepackage{graphicx}%
\usepackage{multirow}%
\usepackage{amsmath,amssymb,amsfonts}%
\usepackage{amsthm}%
\usepackage{mathrsfs}%
\usepackage{fix-cm}%
\usepackage[title]{appendix}%
\usepackage{xcolor}%
\usepackage{textcomp}%
\usepackage{manyfoot}%
\usepackage{booktabs}%
\usepackage{algorithm}%
\usepackage{algorithmicx}%
\usepackage{algpseudocode}%
\usepackage{listings}%
\theoremstyle{thmstyleone}%
\theoremstyle{thmstyletwo}%

\theoremstyle{thmstylethree}%

\usepackage{amsmath,amssymb,amsfonts}
\usepackage{graphicx}
\usepackage{amsmath,amsfonts,amssymb,amsthm}
\usepackage{textcomp}
\usepackage{xspace}
\usepackage{caption}
\usepackage{booktabs}
\usepackage{subcaption}
\usepackage{appendix}
\usepackage{tabularx}
\usepackage[autostyle=true]{csquotes}
\MakeOuterQuote{"}

\usepackage{longtable}
\usepackage{pdflscape}

\usepackage{adjustbox}

\renewcommand{\arraystretch}{1.2}
\usepackage{tikz}
\usetikzlibrary{quantikz2}
\newcommand{\CNOT}{\ensuremath{\mathrm{CNOT}}\xspace}
\newcommand{\CNOTs}{\ensuremath{\mathrm{CNOTs}}\xspace}

\newcommand{\C}{\ensuremath{\mathcal{C}}\xspace}
\newcommand{\XOR}{\ensuremath{\mathrm{XOR}}\xspace}

\usepackage{xcolor}
\newtheorem{problem}{Problem}

\theoremstyle{definition}
\newtheorem{Prop}{Property}
\definecolor{dark_blue}{RGB} {50, 150, 220}
\definecolor{jac_colour}{RGB}{255, 185, 40}
\definecolor{ricca_colour}{RGB}{220, 60, 140}
\definecolor{purple_}{RGB}{100, 72, 150}
\definecolor{done_colour}{RGB}{50, 200, 50}

\begin{document}

\title[Article Title]{AlphaCNOT: Learning CNOT Minimization with
Model-Based Planning}

\author*[1]{\fnm{Jacopo} \sur{Cossio}}\email{cossio.jacopo@spes.uniud.it}

\author[1,2]{\fnm{Daniele} \sur{Lizzio Bosco}}\email{lizziobosco.daniele@spes.uniud.it}

\author[1]{\fnm{Riccardo} \sur{Romanello}}\email{riccardo.romanello@uniud.it}

\author[1]{\fnm{Carla} \sur{Piazza}}\email{carla.piazza@uniud.it}

\author[1]{\fnm{Giuseppe} \sur{Serra}}\email{giuseppe.serra@uniud.it}

\affil[1]{\orgdiv{Department of Mathematics, Computer Science and Physics,} \orgname{University of Udine}, \orgaddress{\city{Udine}, \country{Italy}}}

\affil[2]{ \orgdiv{Department of Biology,} \orgname{University of Naples Federico II}, \orgaddress{\city{Naples}, \country{Italy}}}


\abstract{Quantum circuit optimization is a central task in quantum computing, as current NISQ devices suffer from error propagation that often scales with the number of operations.
Among quantum operations, the CNOT gate is of fundamental importance, being the only two-qubit gate in the universal Clifford+T set.
Minimizing CNOT gates is therefore particularly important, as these gates exhibit higher error rates and they typically introduce additional routing overhead due to hardware constraints.
The problem of CNOT gate minimization has been addressed by heuristic algorithms such as the well-known Patel-Markov-Hayes (PMH) for \textit{linear reversible synthesis} (i.e., CNOT minimization with no topological constraints), and more recently by Reinforcement Learning (RL) based strategies in the more complex case of \textit{topology-aware synthesis}, where each CNOT can act on a subset of qubit pairs.
We introduce \texttt{AlphaCNOT}, an RL framework based on Monte Carlo Tree Search that effectively addresses the CNOT minimization problem by modeling it as a planning problem. In contrast to other RL-based solutions, our method is \textit{model-based}, i.e., it can leverage lookahead search to evaluate future trajectories,  thus finding more efficient sequences of CNOTs.
Our method achieves a reduction of up to 24.62\% in CNOT count compared with the PMH baseline on linear reversible synthesis, and a consistent gate count reduction on different topologies with up to 8 qubits, also with respect to state-of-the-art RL-based solutions. 
Our results suggest the combination of RL with search-based strategies can be applied to different circuit optimization tasks, thus fostering the transition toward the ``quantum utility'' era.}

\keywords{Quantum Compilation, Linear Reversible Synthesis, Reinforcement Learning, AlphaZero, CNOT Synthesis}

\maketitle

\bmhead{Acknowledgements}
    Jacopo Cossio is supported by FSE/FVG PhD Grant on ``Computer Science and Artificial Intelligence'' (CUP G23C25000620008).
    This work has been partially supported by INdAM-GNCS project \emph{Algebra lineare quantistica, state preparation e compilazione di circuiti quantistici} (CUP E53C25002010001) and by the regional project QUASAR-FVG \emph{Calcolo e simulazione quantistica: sviluppo, applicazioni e ricerca in Friuli Venezia Giulia} (CUP G23C25001510002).

\section{Introduction}
\label{sec:introduction}

\begin{figure*}[htb]
    \centering
    \includegraphics[width=0.85\textwidth,
        trim={8 74 245 10}, clip]{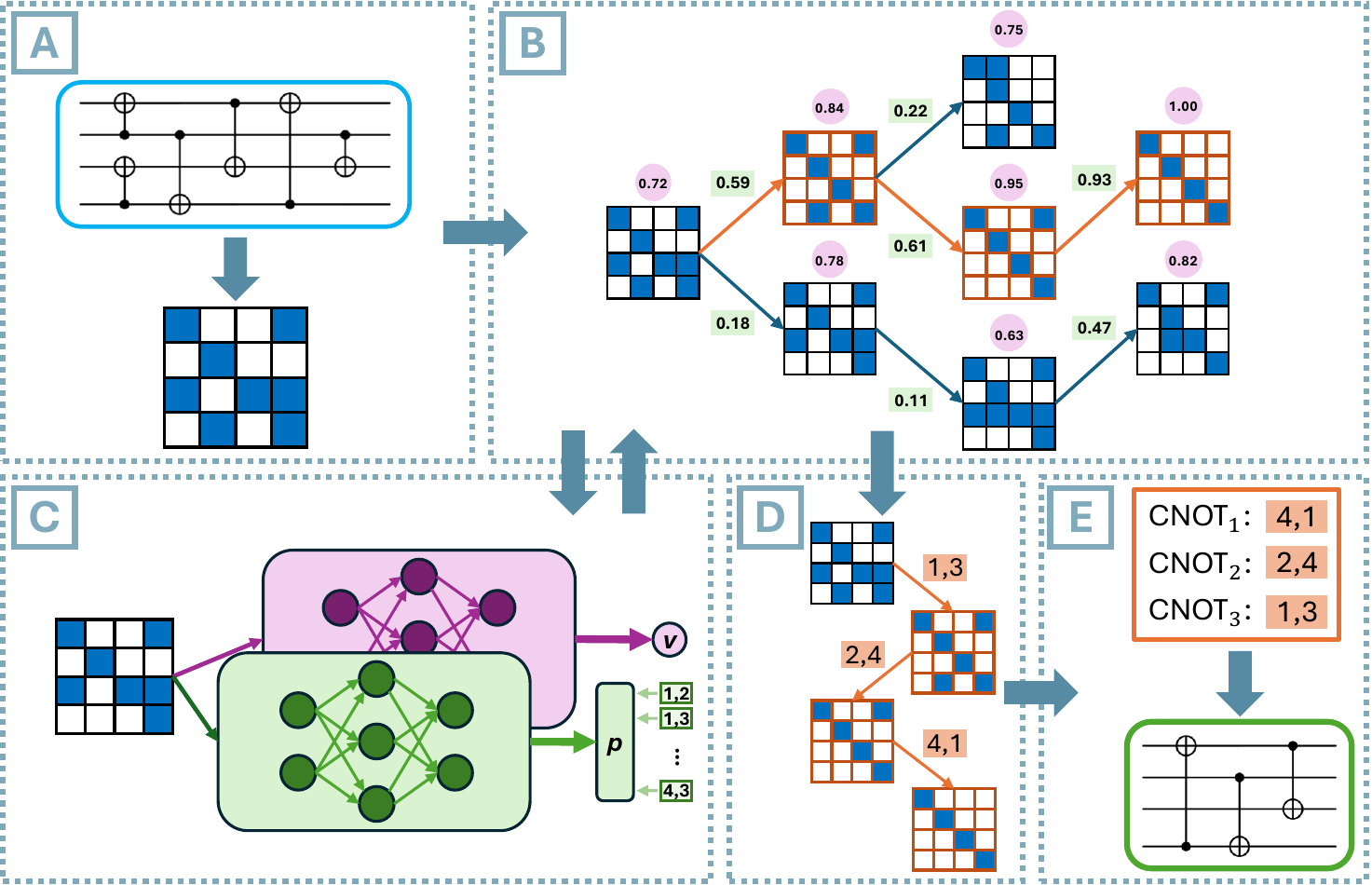}
    \caption{\texttt{AlphaCNOT} Framework. A: The target circuit is encoded into its parity matrix. B: The system dynamics are modeled as a tree, where nodes correspond to different circuits obtainable from the starting parity matrix, based on the values and policies of each node estimated by a pair of Neural Networks (C). D: Eventually, a suitable path from the root node to the identity matrix is identified. E: The \CNOT sequence corresponding to the selected path is reversed to produce the optimized circuit.}
    \label{fig:framework}
\end{figure*}

Significant progress has been observed in quantum computing algorithms, providing interesting applications for several problem classes, including quantum simulation \citep{QSIM0, QSIM15, QSIM1}, optimization \citep{QAOA_2014}, and number-theoretic tasks \citep{shor_algorithms_1994, QuantumDLP}. 

The physical realization of these algorithms requires the definition of abstract circuits that need to be compiled into native gate sets \citep{itoko2020}. Due to the hardware characteristics of current quantum devices, this process introduces several constraints and limitations, including short coherence time and limited qubit connectivity.

Given these hardware limitations, optimizing this compilation process to minimize gate count and circuit depth is a critical requirement for successful execution on current Noisy Intermediate-Scale Quantum (NISQ) hardware \citep{preskill2018}, characterized by a limited number of qubits and a high susceptibility to errors. 

\CNOT gates (Controlled-NOT) are the primary source of qubit interaction in the Clifford+T set, since they are the only available two-qubit gate. At the same time, \CNOTs are significantly more prone to error than single-qubit gates \citep{bruzewicz2019}. 
Therefore, a key point to address is the \CNOT minimization problem, which aims to synthesize a target sequence of \CNOTs into an equivalent formulation with the fewest possible gates.

To achieve these syntheses, two different approaches based on the underlying hardware structure have been developed. The first one is the unconstrained optimization, also called \textit{linear reversible synthesis} \citep{pmh2003}, where every qubit is connected to the other qubits and thus all operations are permitted. On the other hand, the second version is the constrained optimization or \textit{Topology-Aware Synthesis} \citep{kremer2024}, where qubit interactions are limited to those supported by a given quantum device. In the latter case we generally expect longer syntheses due to the limited topology.

Traditional solutions to \CNOT minimization rely on matrix decomposition heuristics, such as Gaussian elimination or the Patel-Markov-Hayes (PMH) algorithm \citep{pmh2003,  schaeffer2014, debrugiere2021}. While computationally efficient, these methods are inherently greedy, i.e., they always opt for the maximum local cost reduction and frequently fail to converge to the global optimum. More recently, Reinforcement Learning (RL) methods have been applied to this domain, often outperforming traditional heuristics, both in the linear reversible synthesis \citep{romanello2025} and in the topology-aware setting \citep{kremer2024}. 
However, these approaches  rely on model-free algorithms, such as Proximal Policy Optimization \citep{schulman2017} (PPO). These algorithms  learn directly from the interactions with the environment and do not exploit an explicit representation of the system dynamics.
While these approaches learn which \CNOT gate to apply at each step following a predefined policy, they only explore a single path at a time, lacking the ability to explicitly plan a synthesis strategy that improves overall solution quality.
Essentially, a model-free agent is comparable to a navigator without a map: it finds the most promising immediate move, but it cannot foresee if this local decision will ultimately result in a dead end or an inefficient path.

To overcome these limitations, we tackle the problem using a tree-based search strategy, which efficiently explores multiple candidate paths and selects the most promising one. We develop \texttt{AlphaCNOT}, sketched in Figure \ref{fig:framework}, a model-based RL approach which relies on AlphaZero, first introduced in \cite{silver2017go, silver2017zero} to solve complex two-player games, like go, chess and shogi. 
The tree structure, central for our method, encapsulates structural information of the problem that facilitates the search for optimal solutions.
In order to increase the expressivity of our reinforcement learning algorithm, we introduce a mixed reward function, which transitions from an initial stage of heuristic-based feedback to a non-informed reward phase.
Our agent addresses the \CNOT minimization problem as a \textit{planning} task by combining a deep neural network with Monte Carlo Tree Search (MCTS) \citep{coulom2007}.

Unlike previous optimization approaches, our method can be applied to both the unconstrained linear reversible synthesis and the topology-constrained variant. 
Our results demonstrate that \texttt{AlphaCNOT} reduces gate counts by up to 24.62\% compared to the well-known PMH algorithm \citep{pmh2003} and consistently outperforms previous model-free RL baselines \citep{romanello2025} and other heuristic algorithms \citep{debrugiere2021, schaeffer2014} on linear reversible synthesis. 
On the more  complex task of topology-aware synthesis, we evaluate different topologies (i.e., connectivity maps) with up to 8 qubits, obtaining again consistent advantages over other methods \citep{kremer2024}. 
In addition, we consider also the permutation-relaxed case, in which the synthesis of the target \CNOT circuit terminates up to a permutation matrix. This variant of our method is referred to as \texttt{PermAlphaCNOT} and it is discussed in Section \ref{SubSection_EXP_permutation}.
Finally, we evaluate our methods on phase polynomial synthesis for QAOA circuits \citep{QAOA_2014} for MaxCut and LABS instances, taking inspiration from \cite{vandaele2022phasesynthesis}, across the 19 topologies used in this work. The comparison between our newly introduced \texttt{AlphaCNOT} and \texttt{PermAlphaCNOT} with state-of-the-art methods for \CNOT synthesis showed better performances of the formers in almost every setting.

To support reproducibility and further research on this topic, we release our source code and pre-trained models \footnote{\url{https://github.com/Jaccos01/AlphaCNOT}}. In our framework, we address the computational cost typically associated with MCTS by providing a highly parallelized implementation based on {JAX} \citep{jax2018github}.

The paper is organized as follows. In Section \ref{Section_BG} we formalize both the unconstrained and constrained \CNOT minimization problems. In Section \ref{Section_RL} we present related work. Our method is described in Section \ref{Section_MCTS}. In Section \ref{Section_EXP} we discuss experimental settings and results. Section \ref{Section_CONCLUSION} closes the paper with final considerations and further work. 

\section{The \CNOT Minimization Problems}\label{Section_BG}
\subsection{\CNOT Circuits}\label{sub3:LRC}
The Controlled-NOT (\CNOT) gate is a fundamental two-qubit operation in quantum computing. Given a control qubit $i$ and a target qubit $j$, the effect of $\CNOT(i,j)$ is flipping the target qubit's value when the control qubit is $\ket{1}$, i.e. the value of $\ket{j}$ after applying $\CNOT(i,j)$ is $ \ket{i\oplus j}$. Its associated unitary, in a 2-qubit setting, is the following
\begin{equation}
    \CNOT = \begin{pmatrix}
        1 & 0 & 0 & 0\\
        0 & 1 & 0 & 0\\
        0 & 0 & 0 & 1\\
        0 & 0 & 1 & 0\\
    \end{pmatrix}.
\end{equation}
Given a $\CNOT (i,j)$ gate in an $n$-qubit system, its associated \textit{elementary parity matrix} $E_{i,j}$ is defined as \begin{equation}
E_{ij} = I_n + \left( \delta_{k j}\,\delta_{l i} \right)_{1 \le k,l \le n},
\end{equation}
i.e., the identity matrix with  an additional $1$ at position $(j, i)$.
Given a Boolean linear reversible circuit $\mathcal{C}$, i.e., a  quantum circuit composed solely of \CNOT gates, $\mathcal{C} = (\CNOT_1, \CNOT_2, \dots, \CNOT_m)$, we can construct its parity matrix $M_{\mathcal{C}}$  by multiplying the elementary parity matrices $E_k$ associated with each gate. Specifically, if the gates are applied in the order $k=1, \dots, m$, the resulting parity matrix is given by $M_\mathcal{C} = E_m  E_{m-1} \dots E_1$, where each product is computed over the field $\mathbb{F}_2$, i.e., the field over $\{0,1\}$.

Note that the parity matrix $M$ of a linear reversible circuit, hereafter called a \CNOT circuit, can also be defined as the only Boolean matrix satisfying
\begin{equation}
    y_i = \bigoplus_{j=1}^n M_{ij}x_j,
\end{equation}
where $y_i$ is the output value of the $i$-th wire of the circuit, while $x_j$ is the input value of the $j$-th wire.
An example of a \CNOT circuit with the corresponding parity matrix is provided in Figure \ref{fig:cnot&parity}. 

\begin{figure*}[h]
    \centering
    \begin{tabular}{c@{\hspace{2cm}}c} 
        $\mathcal{C}:$
        \begin{tikzcd}[row sep=0.25cm]
        \lstick{$x_1$} & \ctrl{1} & \targ{} & \ctrl{5} & \qw & \qw & \rstick{$x_2$} \\
        \lstick{$x_2$} & \targ{}  & \ctrl{-1} & \qw & \targ{} & \qw & \rstick{$x_1 \oplus x_3 \oplus x_6$} \\
        \lstick{$x_3$} & \ctrl{1} & \ctrl{3} & \qw & \qw & \qw & \rstick{$x_3$} \\
        \lstick{$x_4$} & \targ{}  & \qw & \qw & \qw & \qw & \rstick{$x_3 \oplus x_4$} \\
        \lstick{$x_5$} & \targ{}  & \qw & \qw & \qw & \qw & \rstick{$x_5 \oplus x_6$} \\
        \lstick{$x_6$} & \ctrl{-1} & \targ{} & \targ{} & \ctrl{-4} & \qw & \rstick{$x_2 \oplus x_3 \oplus x_6$}
        \end{tikzcd}
        
        & 
        
        $M_\C = 
        \begin{pmatrix}
        0 & 1 & 0 & 0 & 0 & 0 \\
        1 & 0 & 1 & 0 & 0 & 1 \\
        0 & 0 & 1 & 0 & 0 & 0 \\
        0 & 0 & 1 & 1 & 0 & 0 \\
        0 & 0 & 0 & 0 & 1 & 1 \\
        0 & 1 & 1 & 0 & 0 & 1
        \end{pmatrix}$
        
    \end{tabular}
    \caption{Parity matrix $M_{\C}$ of a given circuit $\C$.}
    \label{fig:cnot&parity}
\end{figure*}
On the other hand, any invertible Boolean matrix $M \in GL(n, \mathbb{F}_2)$ can be synthesized into a linear reversible circuit composed solely of \CNOT gates. 

\begin{Prop}\label{prop:CNOT_XOR_equivalence}
Given a \CNOT circuit $\mathcal{C}$ and its corresponding parity matrix $M_\mathcal{C}$, performing a \CNOT with control $i$ and target $j$ on \C is equivalent to applying a \XOR operation between rows $i$ and $j$ to the $j$-th row on the parity matrix $M$, i.e., $R_j \leftarrow R_j \oplus R_i$.
\end{Prop}
It is important to emphasize that the mapping between representations of \CNOT circuits and parity matrices is not injective. Different sequences of \CNOT gates can represent the same quantum circuit, in which case we call them \emph{equivalent} as they result in the same quantum operation and in the same parity matrix $M$. Figure \ref{fig:equivalent_circuits} shows two equivalent representations of the same circuit, 

\begin{figure}[h]
    \centering
    \begin{minipage}{0.40\linewidth}
        \centering
        % $\C_1:$
        \begin{tikzcd}[row sep=0.2cm, column sep=0.3cm]
        \lstick{$x_1$} & \ctrl{2} & \targ{} & \qw & \qw & \targ{} & \rstick{$x_1$} \\
        \lstick{$x_2$} & \qw  & \ctrl{-1} & \targ{} & \ctrl{1} & \qw & \rstick{$x_1\oplus x_2\oplus x_3$} \\
        \lstick{$x_3$} & \targ{} & \qw & \ctrl{-1} & \targ{} & \ctrl{-2} & \rstick{$x_2$}\\
        \end{tikzcd}
    \end{minipage}
    \hfill
    \begin{minipage}{0.40\linewidth}
        \centering
        % $\C_2:$
        \begin{tikzcd}[row sep=0.25cm]
        \lstick{$x_1$} & \qw & \qw &  \ctrl{1} & \rstick{$x_1$} \\
        \lstick{$x_2$} & \targ{}  & \ctrl{1} & \targ{} & \rstick{$x_1\oplus x_2\oplus x_3$} \\
        \lstick{$x_3$} & \ctrl{-1} & \targ{} & \qw & \rstick{$x_2$} \\
        \end{tikzcd}
    \end{minipage}
    \caption{Two equivalent \CNOT circuits.
    }
    \label{fig:equivalent_circuits}
\end{figure}
both of them associated with the parity matrix
\begin{equation*}
M = 
        \begin{pmatrix}
        1 & 0 & 0  \\
        1 & 1 & 1 \\
        0 & 1 & 0  \\
        \end{pmatrix}.
\end{equation*}
Since each \CNOT circuit \(\mathcal{C}\) can be mapped to a parity matrix \(M_{\mathcal{C}}\), and any \CNOT circuit can be expressed as a concatenation of \CNOT gates \(\mathcal{C}= C_1 \cdots C_n\) with \(M_{\mathcal{C}} = M_n \cdots M_1\) for the corresponding parity matrices \(M_i\), in what follows we will use the circuit and its parity matrix interchangeably whenever no ambiguity arises. 

\subsection{\CNOT Minimization on NISQ Hardware}

On current and near-future Noisy Intermediate-Scale Quantum (NISQ) hardware, two-qubit gates introduce many challenges, as they are usually more error-prone compared to single-qubit gates \citep{preskill2018}.
This motivates the introduction of the following problem:
\begin{problem}[\CNOT Minimization]\label{Prob1}
    Given a \CNOT circuit $\mathcal{C}$ on $n$ qubits, find a decomposition $\mathcal{C}=C_1\dots C_k$, where each $C_i$ is a \CNOT gate, of minimal length $k$.
\end{problem}
As a consequence of Property \ref{prop:CNOT_XOR_equivalence}, Problem \ref{Prob1} is equivalent to finding the shortest sequence of \XOR operations between rows that reduces \(M_\mathcal{C}\) to the identity matrix. 
It has recently been proved to be NP-hard: in particular,  \cite{Acuaviva_2026} and  \cite{Li_2026} provided two independent proofs showing that the decisional problem of finding a \CNOT synthesis with less than $k$ \CNOTs is NP-complete.
Theoretically, it has been shown that $\Theta(n^2 / \log n)$ \CNOT gates are required in the worst case \citep{pmh2003}. Consequently, Problem \ref{Prob1} has been addressed by various heuristics, which provide polynomial-time syntheses without aiming at reaching the exact minimum for every instance. The most important ones, including Patel-Markov-Hayes (PMH) Algorithm \citep{pmh2003}, AECM and MCG \citep{schaeffer2014}, and  GreedyGE \citep{debrugiere2021}, are discussed in the following section.

Note that Problem \ref{Prob1} is meaningful as long as the considered hardware exhibits  full connectivity (i.e., each qubit can directly interact with all other qubits).
In many quantum computing paradigms, including  superconducting quantum devices \citep{arute2019, jurcevic2021},  the hardware connectivity is restricted, i.e., not all qubit pairs can interact directly.
More in detail, each hardware device has an associated \textit{topology}, corresponding to the set of pairs of qubits that can interact. In this regard, the operation $\CNOT(i,j)$ can be executed natively only when the pair $(i,j)$ belongs to the topology of the device. When this is not the case, interaction typically requires a \textit{SWAP} operation, which consumes three \CNOT gates.
This motivates the following problem variant.
\begin{problem}[Topology Aware \CNOT Minimization]\label{Prob2}
        Given a \CNOT circuit $\mathcal{C}$ and a topology $\mathcal{T}$ on $n$ qubits, find a decomposition $\mathcal{C}=C_1\dots C_k$  of minimal length $k$, where each $C_i$ is a \CNOT gate between a pair of qubits in $\mathcal{T}$.
\end{problem}

This variant introduces additional constraints, making many heuristics developed for Problem \ref{Prob1} (e.g., PMH) not applicable. 
Similarly to Problem \ref{Prob1}, Problem \ref{Prob2} is NP-hard for an arbitrary topology $\mathcal{T}$ \citep{jiang2020}.
Thus, we cannot rely on exact methods and must instead use alternative greedy solvers as the input dimensions increase.

\subsection{Phase Polynomials and Trailing CNOT Circuits}\label{sub3:phasepolynomials}

An important class of circuits in which \CNOT synthesis naturally arises is the one of \emph{\CNOT-phase circuits}, i.e., circuits composed solely of \CNOT and $R_Z$ gates~\citep{amy2018}. Following~\cite{vandaele2022phasesynthesis}, the action of such a circuit $U$ on $n$ qubits can be described, for every computational basis state $\ket{x}$ with $x\in\mathbb{F}_2^n$, as
\begin{equation}
    U\ket{x}=e^{\mathrm{i}p(x)}\ket{g(x)},
    \qquad
    p(x)=\sum_{k=1}^{m}\theta_k f_k(x),
    \label{eq:ricca_phase_polynomial}
\end{equation}
where $g:\mathbb{F}_2^n\to\mathbb{F}_2^n$ is a linear reversible function, i.e., $g(x)=M_g x$ for some parity matrix $M_g\in GL(n,\mathbb{F}_2)$, and $p$ is the \emph{phase polynomial} of the circuit. Each term of $p$ is a linear Boolean function $f_k(x)=y^{(k)}\cdot x=\bigoplus_{j=1}^{n}y^{(k)}_j x_j$ (treated as an integer in $\{0,1\}$ inside the sum), weighted by an angle $\theta_k$. The vectors $y^{(k)}\in\mathbb{F}_2^n$ are called the \emph{parities} of $p$ and, arranged as columns, they form the so-called \emph{parity table}~\citep{vandaele2022phasesynthesis}. 

Synthesizing a \CNOT-phase circuit requires synthesizing both its phase polynomial and its linear reversible function $g$~\citep{vandaele2022phasesynthesis}. Each parity $y^{(k)}$ must appear on some wire to apply the corresponding $R_Z(\theta_k)$ gate, since $R_Z$ gates only act on the phase. For this reason, the synthesis of $p$ reduces to finding a \CNOT circuit in which every parity of the table appears at least once, the so-called \emph{parity network}~\citep{amy2018}.

The most straightforward way to build such a network is to treat each term of $p$ as a \emph{phase gadget}, i.e., as the multi-qubit rotation
\begin{equation}
    R_{Z,S}(\theta)=\exp\left(-\frac{\mathrm{i}\theta}{2}\prod_{j\in S}Z_j\right),
    \label{eq:ricca_phase_gadget}
\end{equation}
where $S=\{j : y_j=1\}$ is the support of the corresponding parity $y$. Up to a global phase, $R_{Z,S}(\theta)$ multiplies $\ket{x}$ by $e^{\mathrm{i}\theta(y\cdot x)}$. A phase gadget is implemented by \emph{computing} the parity of the wires in $S$ with a ladder of \CNOTs, applying a single $R_Z(\theta)$ on the target wire, and then \emph{uncomputing} the parity with the reversed ladder~\citep{meijer2025phasepolynomial}. However, synthesizing every gadget on its own wastes many \CNOTs, as shown in Figure~\ref{fig:ricca_pp_naive}.

\begin{figure}[htbp]
    \centering

    % === (a) Naive synthesis ===
    \begin{subfigure}[t]{0.49\textwidth}
        \centering
        \resizebox{\linewidth}{!}{%
        \begin{quantikz}[row sep=0.25cm, column sep=0.3cm]
            \lstick{$x_1$} & \ctrl{1}\gategroup[3,steps=3,style={dashed,rounded corners,inner xsep=2pt},background,label style={label position=above,anchor=south,yshift=-0.1cm}]{{\scriptsize gadget $\theta_1$}} & & \ctrl{1} & \ctrl{1}\gategroup[3,steps=5,style={dashed,rounded corners,inner xsep=2pt},background,label style={label position=above,anchor=south,yshift=-0.1cm}]{{\scriptsize gadget $\theta_2$}} & & & & \ctrl{1} & \rstick{$x_1$} \\
            \lstick{$x_2$} & \targ{} & \gate{R_Z(\theta_1)} & \targ{} & \targ{} & \ctrl{1} & & \ctrl{1} & \targ{} & \rstick{$x_2$} \\
            \lstick{$x_3$} & & & & & \targ{} & \gate{R_Z(\theta_2)} & \targ{} & & \rstick{$x_3$}
        \end{quantikz}%
        }
        \caption{Naive synthesis: each gadget computes, rotates, and uncomputes its own parity (6 \CNOTs).}
        \label{fig:ricca_pp_naive}
    \end{subfigure}
    \hfill
    % === (b) Parity reuse ===
    \begin{subfigure}[t]{0.49\textwidth}
        \centering
        \resizebox{\linewidth}{!}{%
        \begin{quantikz}[row sep=0.25cm, column sep=0.3cm]
            \lstick{$x_1$} & \ctrl{1}\gategroup[3,steps=4,style={dashed,rounded corners,fill=gray!12,inner xsep=2pt},background,label style={label position=above,anchor=south,yshift=-0.1cm}]{$U_{\mathrm{pre}}$} & & & & \gategroup[3,steps=2,style={dashed,rounded corners,fill=blue!12,inner xsep=2pt},background,label style={label position=above,anchor=south,yshift=-0.1cm}]{$U_{\mathrm{tail}}$} & \ctrl{1} & \rstick{$x_1$} \\
            \lstick{$x_2$} & \targ{} & \gate{R_Z(\theta_1)} & \ctrl{1} & & \ctrl{1} & \targ{} & \rstick{$x_2$} \\
            \lstick{$x_3$} & & & \targ{} & \gate{R_Z(\theta_2)} & \targ{} & & \rstick{$x_3$}
        \end{quantikz}%
        }
        \caption{Parity reuse: the uncomputing \CNOTs are pushed to the end, where they form the trailing \CNOT circuit (4 \CNOTs).}
        \label{fig:ricca_pp_tail}
    \end{subfigure}

    \caption{Two syntheses of the phase polynomial
    $p(x)=\theta_1(x_1\oplus x_2)+\theta_2(x_1\oplus x_2\oplus x_3)$,
    with $g(x)=x$, on a 3-qubit line.
    In (b), the parity $x_1\oplus x_2$ computed for $\theta_1$
    is reused to obtain $x_1\oplus x_2\oplus x_3$ with a single
    \CNOT, and all the CNOTs following the last rotation are
    collected in $U_{\mathrm{tail}}$.
    Adapted from ~\cite{meijer2025phasepolynomial}.}
    \label{fig:ricca_phase_poly}
\end{figure}

For this reason, standard phase-polynomial synthesis algorithms, such as GraySynth~\citep{amy2018}, ParitySynth~\citep{vandaele2022phasesynthesis}, and their architecture-aware variants~\citep{meijer2025phasepolynomial}, do not uncompute each parity right after the corresponding rotation. Instead, they compute the parity, apply the rotation, and push the undoing \CNOTs towards the end of the circuit, updating the parity table accordingly. 
% In this way, the parities currently stored on the wires can be reused during the computation: the next gadget is built on top of them, often requiring just a few additional \CNOTs (e.g., by visiting the parities in a Gray-code order~\citep{amy2018}). 
Figure~\ref{fig:ricca_pp_tail} shows this idea on a small example, where the parity $x_1\oplus x_2$, computed for $\theta_1$, is reused to obtain $x_1\oplus x_2\oplus x_3$ with a single \CNOT.

Nonetheless, once the last rotation has been applied, the wires are left in some linear state $M_{\mathrm{pre}}x$, which in general differs from the desired output $g(x)$. Thus, at the end of the synthesis a \CNOT-only circuit emerges, the so-called \emph{trailing \CNOT circuit}~\citep{meijer2025phasepolynomial}, which collects all the postponed uncomputations together with the linear reversible function $g$. More precisely, denoting by $U_{\mathrm{pre}}$ the synthesized prefix (containing all the rotations) and by $U_{\mathrm{tail}}$ the trailing circuit, we have
\begin{equation}
    U=U_{\mathrm{tail}}\,U_{\mathrm{pre}},
    \qquad
    M_{\mathrm{tail}}=M_g\,M_{\mathrm{pre}}^{-1},
    \label{eq:ricca_prefix_tail}
\end{equation}
where $M_{\mathrm{tail}}$ is the parity matrix of $U_{\mathrm{tail}}$. Since the trailing \CNOTs are simply accumulated during the synthesis, they are typically left unoptimized~\citep{meijer2025phasepolynomial}. However, $U_{\mathrm{tail}}$ contains no rotation: it can be replaced by any \CNOT circuit with parity matrix $M_{\mathrm{tail}}$ that respects the hardware topology, without altering the overall operation. In other words, its re-synthesis is exactly an instance of Problem~\ref{Prob1} (or of Problem~\ref{Prob2}, in the topology-aware case), which makes trailing \CNOT circuits a natural, application-driven benchmark for our method.

\section{Related Work}\label{Section_RL}

The problem of minimizing \CNOT gates has been addressed in the literature with different methodologies, including exact methods \citep{meuli2018, piazza2023} and heuristic algorithms  \citep{pmh2003,schaeffer2014, debrugiere2021}.
More recently, Reinforcement-Learning (RL) based solutions have been proposed \citep{kremer2024, romanello2025}, due to the ability of RL model to explore efficiently complex solution spaces. 
These three methodologies are discussed respectively in Sections \ref{sub:RL-EM}, \ref{sub:RL-GH} and \ref{sub:RL-RL}.

\subsection{Exact Methods}\label{sub:RL-EM}
By exact methods, we refer to approaches that model the problem through a logical encoding to find optimal solutions.

A first approach to model \CNOT-minimization in the linear reversible circuit setting (i.e., Problem \ref{Prob1}) is given in \cite{meuli2018}. In this work, the authors develop an algorithm based on the Boolean Satisfiability  \citep{cook1971} (SAT) problem, where each variable is represented by a Boolean value, to minimize the \CNOT count in circuits  composed of \CNOT and T gates. By representing the circuit through \textit{phase polynomials}, they build a decision problem to determine whether a circuit with exactly $k$ \CNOT gates exists for a given initial representation. Starting with a given $k_0$, if the solver determines the problem as unsatisfiable, the bound $k$ is increased. 
Eventually, the SAT solver finds the minimum \CNOT count required to implement the target transformation, providing exact optimal solutions.

Similarly, in \cite{piazza2023}, the authors propose an Answer Set Programming \citep{gebser2013} (ASP) encoding to address exactly Problem \ref{Prob1}. In this work, the task is modeled as finding a sequence of $\CNOT_{ij}$ to map the initial parity matrix $M_\C$ into the identity matrix $I_n$. At the end of the process, they reverse the order of the sequence found to construct $M_\C$ starting from the identity.

These approaches are always guaranteed to obtain the minimal \CNOT count. On the other hand, due to their exponential time complexity they do not scale (e.g., in \cite{piazza2023} it is shown that the method does not find a solution in reasonable time for more than $7$ qubits). 

\subsection{Greedy and Heuristic Algorithms}\label{sub:RL-GH}

Heuristic algorithms provide quick and scalable solutions, at the cost of optimality.
The first heuristic algorithm to address linear reversible circuit synthesis is the Patel-Markov-Hayes (PMH) algorithm \citep{pmh2003}.  In this influential work, the authors introduce a variant of classical Gaussian reduction for synthesizing \CNOT circuits. The core idea of their algorithm consists of grouping together $m<n$ columns and working separately on every section, eliminating duplicate sub-rows and thus reducing the matrix to an upper triangular matrix. 
Overall, the total number $C(n)$ of \CNOT required, while depending on the choice of $m$, can be upper bounded by $n^2/\left(\alpha\log_2(n)\right)$, for a given $\alpha\in(0,1)$.

After the introduction of PMH, two other heuristic  algorithms have been proposed in \cite{schaeffer2014}: the Alternating Elimination with Cost Minimization method (AECM) and the Multiple \CNOT Gate method (MCG). Compared to PMH, both AECM and MCG can reduce rows or columns in a nondeterministic fashion, providing a major cost reduction in terms of number of  \CNOT used. 

These two algorithms have been shown to perform better than PMH under certain conditions (e.g., number of qubits $n$). However, they are expected to perform worse than PMH for large number of qubits (i.e., $n\geq 64$).

A more recent hybrid approach is represented by the GreedyGE algorithm introduced in \cite{debrugiere2021}. 
The parity matrix $M_{\C}$ of a \CNOT circuit $\C$ is first decomposed into the product of two triangular matrices, $L$ and $U$, through the LU factorization. 
This allows the algorithm to restrict the search space and apply a greedy row-reduction strategy to $L$ and $U$ independently.
This algorithm was developed to overcome the limitations of both standard Gaussian elimination, that often gives sub-optimal solutions, and purely greedy search methods, which suffer from poor scalability. 

Another relevant approach is RowCol, introduced by \cite{wu2023rowcol}, which addresses \CNOT circuit synthesis under limited qubit connectivity (see Problem \ref{Prob2}). This method exploits \textit{Steiner trees}, that are minimum-cost trees connecting a specified set of vertices in a graph, possibly through additional intermediate vertices. In short, RowCol iteratively eliminates rows and columns of the parity matrix $M_\mathcal{C}$ of a given \CNOT circuit $\mathcal{C}$, using Steiner trees to ensure that all operations respect the hardware connectivity constraints.

\subsection{Reinforcement Learning}\label{sub:RL-RL}
An alternative approach to tackle the \CNOT minimization problem involves the application of reinforcement learning (RL) \citep{sutton1998, sutton1999}  models. RL is a branch of artificial intelligence, based on an autonomous agent which learns how to make sequences of decisions through interaction with the environment. Every action the agent performs is evaluated by a reward function, reporting a positive or  negative response to the agent. The goal of the agent is to reach a final state, starting from an initial one, maximizing the total reward throughout the process.

\textcolor{black}{Recently, RL has been employed for quantum circuit synthesis \citep{wang2024quantumcompilingreinforcementlearning, kremer2025optimizingnoncliffordcountunitarysynthesis, romanello2025, dubal2025pauli, rietsch2024gumbel, acampora2025PPO}, as well as for related tasks such as state preparation \citep{zen2025stateprep,doherty2026stateprep, RLxZXC, alphatensor_quantum}, hardware-aware compilation \citep{QCGame},  pre-synthesis optimization \citep{qpresyn}, and selection of circuit optimization passes \citep{mills2026optpass}.
}

With regard to \CNOT minimization, the first work addressing the topology-aware synthesis (i.e., Problem \ref{Prob2}) is \cite{kremer2024}.
In this work, the authors consider the problem of circuit synthesis as a \emph{sequential decision process}: at every time step $t$, a chosen \CNOT gate $g_t$ is applied to the current operator $O_t$ resulting in another invertible operator $O_{t+1}$. Starting from the initial matrix $O_0$, the process is iteratively repeated for a given number of steps $T$ until the identity operator $O_T = I$ is obtained.
They exploit \textit{curriculum learning} \citep{bengio2009}, a strategy in which the agent is asked to solve iteratively more complex tasks, and a sparse reward function (i.e., the agent is rewarded only when it reaches the target state) to enhance the agent's training.

With regard to  Problem \ref{Prob2}, they show that their proposed agent, based on Proximal Policy Optimization \citep{schulman2017} (PPO), obtains better results than a PMH variant for topology-aware synthesis (i.e., PMH on the unconstrained setting, followed by SABRE \citep{zou2024} for routing) on a large selection of topologies.

\textcolor{black}{A related PPO-based approach addresses topology-aware linear reversible synthesis, focusing on CX depth reduction \citep{acampora2025PPO}.}

More recently, in \cite{romanello2025} the authors address the unconstrained version by employing a PPO agent trained on a fixed dimension ($n=8$ qubits), and then evaluate on tasks of different sizes. In particular, to solve instances of smaller sizes $m<n$ they embed the matrix representation in one of size $n$. For higher dimensions, they first apply a PMH-based reduction to ``fix'' the first $m-n$ columns, and then employ the RL agent to solve the remaining $n\times n$ submatrix.
In contrast to \cite{kremer2024}, the authors implement an informed reward function based on the Hamming distance. By combining this approach with a curriculum-learning setting where the agent observes ``easier'' instances before addressing more general, complex ones, they obtain a consistent reduction compared to PMH.

Despite their promising results, the approaches implemented in \cite{kremer2024, acampora2025PPO}, and \cite{romanello2025} do not exploit the known algebraic dynamic of the problem.

\textcolor{black}{RL combined with tree search has also been explored for more general unitary synthesis, using Gumbel AlphaZero \citep{rietsch2024gumbel} and AlphaZero-inspired planning \citep{valcarce2025alphazero}.}

More recently, learning-based methods have also been deployed for
Clifford circuit synthesis, which generalizes the CNOT synthesis
problem considered here. AlphaClifford~\citep{LizzioBosco_AlphaClifford}, \textcolor{black}{ inspired by the AlphaCNOT framework presented here, extends model-based MCTS planning to both unconstrained and topology-aware Clifford circuit synthesis.} In contrast,
\cite{Richie_2026} exploits a model-free approach based on an equivariant
neural network, designed to generalize across different cardinalities. The results shown in these works  provide complementary evidence for the potential of learning-based approaches to quantum optimization.

In this work, we focus specifically on CNOT minimization using a model-based RL framework based on Monte Carlo Tree Search.

\section{\texttt{AlphaCNOT}: Learning \CNOT Minimization}\label{Section_MCTS}
Among various reinforcement learning techniques, methods such as Proximal Policy Optimization \citep{schulman2017} (PPO) are referred to as \emph{model-free}, since they learn only through interactions with the environment without constructing an explicit model of its dynamics. 
These methods operate by analyzing sequences of state–action–reward transitions and updating the policy parameters through gradient-based optimization. Since they do not rely on a model of the environment, they are particularly suitable for problems where the system dynamics are unknown (like ``CartPole'' \citep{barto1983}) or difficult to model (like robot simulations) \citep{levine2018}.
On the other hand, \emph{model-based} algorithms build a model of the environment, which is then used to simulate possible scenarios for future actions \citep{sutton1998}. In complex combinatorial spaces, such as the space of \CNOT circuits (i.e., the set of  invertible Boolean matrices), their ability to explore efficiently and in a structured way allows them to achieve better and more scalable results.

Our framework, shown in Figure \ref{fig:framework}, is organized as follows. First, the initial circuit is converted to its parity matrix (A) which is embedded in the tree-shaped network (B). Within the search tree, each node corresponds to a specific matrix, while each (directed) edge represents a \CNOT gate application that transforms one matrix into another. The exploration criterion is based on the values and policies provided by the corresponding neural networks (C), which evaluate each matrix of the tree. At the end of the process, the chosen path defines a sequence of \CNOT moves (D) that define a new equivalent circuit (E).

\subsection{Problem Modeling}
In what follows, we specifically aim at reducing the initial state to the identity matrix. However, the treatment is analogous when the target is a general permutation matrix, as proposed in \cite{meijer2023permrowcol}. The latter variant of Problem \ref{Prob1} is discussed in Section \ref{SubSection_EXP_permutation}.
The search space of our problem, represented by the set of invertible Boolean matrices, is explored through a tree-based approach. Given a fixed problem dimension $n$ (corresponding to the number of  qubits), a target \CNOT circuit $\C$ represented by its parity matrix $M_{\C}\in \mathcal{M}_n(\mathbb{F}_2)$, and possibly a topology $\mathcal T$, we aim to find a composition of \CNOTs equivalent to $\C$. To this aim, we build a tree in the following way:
\begin{itemize}
    \item The root node represents the target invertible matrix $M_\C$;
    \item Each node $N$ is labeled by a non-identity matrix $M$ and it has a child for each possible $\CNOT_h$ in $\mathcal{T}$ (in the unconstrained setting, we can suppose $\mathcal{T}$ to be the fully-connected graph). The child node is labeled by the matrix obtained by applying $\CNOT_h$ to $M$;
    \item The terminal nodes (i.e., the leaves) are labeled by the identity $I_n$.
\end{itemize}

As described in Section \ref{Section_BG}, each $\CNOT_h$ is defined by a control qubit $i_h$ and a target qubit $j_h$. Therefore, in the unconstrained setting, each node $N$ has $\Theta(n^2)$ children.
Note that a path from $M_\C$ to $I_n$ describes a sequence of \CNOTs $\CNOT_1,\dots,\CNOT_k$, such that $(E_1\dots E_k) M_C = I_n$, where each $E_i$ is the elementary parity matrix corresponding to $\CNOT_i$, as described in Section \ref{Section_BG}. For this reason, $\CNOT_k\dots \CNOT_1$ is a decomposition of the target circuit $\C$. In other words, the shortest path from $M_\C$ to $I_n$ is a solution of the \CNOT  minimization problem. See Figure \ref{fig:four_panel_pdf} (left-most panel) for a representation of the selection of a suitable path through the search tree.

\subsection{ The \texttt{AlphaCNOT} Framework}

\begin{figure*}[htb]
    \centering
    \includegraphics[page=1, width=0.21\textwidth,
        trim={0 22 0 0}, clip]{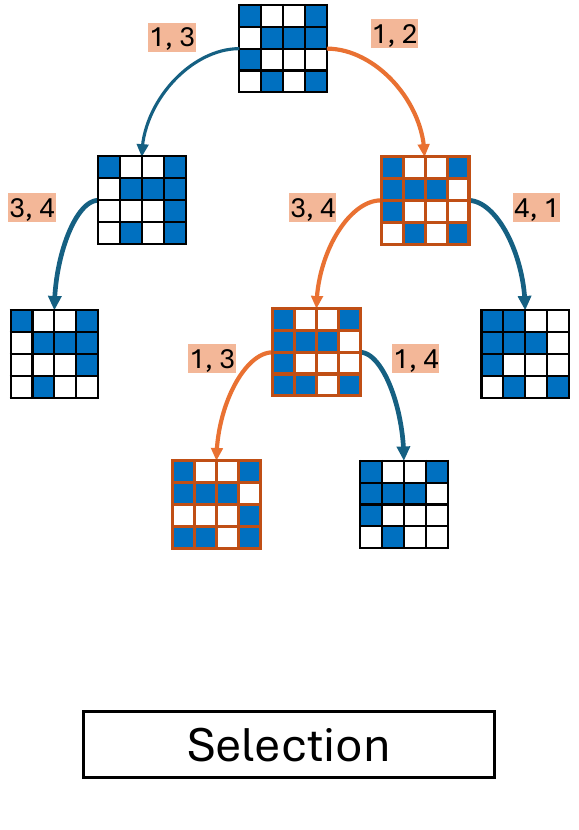}
    \hfill
    \includegraphics[page=2, width=0.21\textwidth,
        trim={0 22 0 0}, clip]{images_new/schemes_prod4.pdf}
    \hfill
    \includegraphics[page=3, width=0.21\textwidth,
       trim={0 22 0 0}, clip]{images_new/schemes_prod4.pdf}
    \hfill
    \includegraphics[page=4, width=0.21\textwidth,
        trim={0 22 0 0}, clip]{images_new/schemes_prod4.pdf}
    \caption{Four-panel visualization of MCTS paradigm. Selection: starting from the root node, which represents the parity matrix of the initial circuit, a path is selected following sequential halving for the root node and Gumbel AlphaZero search for internal actions. Expansion: whenever a non-terminal leaf is reached, the tree is expanded with a new node. Simulation: following the value and policy networks, a thorough rollout is played from the newly expanded node. Backpropagation: all the values and visit statistics are backward updated from the terminal state to the root.}
    \label{fig:four_panel_pdf}
\end{figure*}

The model-based method introduced in this paper, \texttt{AlphaCNOT}, adopts a paradigm that integrates deep learning with stochastic search via Monte Carlo Tree Search (MCTS) \citep{coulom2007}.
The latter is a heuristic search algorithm that incrementally builds a search tree via Monte Carlo rollouts, so as to approximate action-value functions. Since the exploration tree becomes exponentially large as training proceeds, Monte Carlo Tree Search addresses this challenge by expanding branches of the tree that \emph{promisingly} lead to higher expected returns. The MCTS paradigm is composed of four main phases:
\begin{enumerate}
    \item \textbf{Selection:} starting from the initial matrix $M_C$, the tree is visited by iteratively choosing \CNOT applications according to a selection policy (Gumbel AlphaZero search \citep{danihelka2022gumbel}) until a leaf node is reached. This policy balances exploration of less-visited nodes and exploitation of nodes with high estimated value;

    \item \textbf{Expansion:} if the selected node is non-terminal and has unvisited actions, the tree is expanded by applying a new \CNOT move, \emph{de facto} introducing a new child node. A node is terminal either if it is labeled by the identity matrix or it satisfies some other properties described in the environment (e.g., depending on the height of the node). 
    Whenever a terminal node is reached, the iteration  goes directly to the Backpropagation step;

    \item \textbf{Simulation:} from the newly expanded node, a rollout (or playout) is performed by simulating a sequence of \CNOTs according to a policy until a terminal node is reached and the outcome (or reward) can be calculated; 

    \item \textbf{Backpropagation:} the result of the simulation is propagated back through the selected path, updating the parameters (visit counts and value estimates) of all nodes encountered during the selection phase.
\end{enumerate}
A representation of the four steps is provided in Figure \ref{fig:four_panel_pdf}.

The training process in MCTS proceeds iteratively through a loop of steps $1-4$. Each iteration starts at the node representing the initial matrix, and follows a selection policy to visit the tree by choosing actions following the tree policy. This descent continues until a leaf node is reached, corresponding either to an unexpanded state or to a terminal state.
If the selected node is non-terminal and not fully expanded, the expansion phase comes into play: the tree is extended by adding one or more child nodes corresponding to previously unvisited actions. From the newly expanded node, a rollout is then performed by simulating a sequence of actions according to a given default policy straight to a terminal node.
Once the outcome of the simulation is available, the algorithm performs the backpropagation phase, during which the obtained reward is propagated backward along the path from the expanded node to the root. During this process, each visited node updates its visit counts and value estimations, which will be used to guide future selections. This cycle is repeated either for a fixed number of iterations or until a predefined time limit.

Since a complete exploration of the graph is computationally infeasible, the general algorithm is enriched with two neural networks, called \textit{policy network ($p$)} and \textit{value network ($v$)}. The policy network determines a prior probability distribution over the applicable \CNOT gates, providing, at each step, a statistical guide that prioritizes gates more likely to lead to the target identity matrix. The value network, on the other hand, provides a continuous estimation of the state's quality: it informs the algorithm about the potential outcome of a specific branch, without the need for exhaustive and expensive rollouts. 

\subsection{Policy and Value Networks}

Both networks share a common architecture based on a Residual MLP (9 layers with 256 neurons each). Our choice is completely arbitrary, even though the framework allows for any neural architecture depending on the task complexity.

In more detail, each binary matrix $M$ is initially flattened into a vector and then mapped to the first hidden layer through a linear transformation. We implemented skip connections every two layers, for a total of three skip connections throughout the net. These residual blocks facilitate the training process both by mitigating the vanishing gradient problem, and by allowing the network to propagate information across layers without losing the structural features of the input matrix.
The policy head outputs a probability distribution $p$ over the
possible \CNOT actions, e.g., $n(n-1)$ in the general all-to-all topology. 
Conversely, the value head outputs a single number $v$, representing the expected return of the current matrix configuration, i.e., how far the state is from a leaf.

\subsection{Reward Function Design}\label{sub:Reward Design}
When modeling the environment's reward function, it is important to note that most RL algorithms struggle when the reward space is \textit{sparse}, e.g., when most episodes have no positive reward. For this reason, a simple reward of the form 
\begin{equation}
\label{eq:reward_simple}
    \text{Reward}_S(M_t) = \begin{cases}
        0 & \text{if } M_t\ne I\\
        1 & \text{otherwise,}
    \end{cases}
\end{equation}
despite being natural for this problem, would hinder the learning process, as a non-trained model would have no chance to solve any initial episode, leading to constant reward. A possible solution to this limitation is the use of curriculum learning \citep{bengio2009}. However, it requires defining a distribution of easy and hard instances, introducing additional complexity.

To address this limitation, we used an \textit{informed reward}: in particular, we reward each step $t$ based on the Hamming distance Hamming$(M_t, I)$. In this way we are able to "suggest" a possible direction to the agent to reach the identity matrix, solving the episode. The risk of this approach lies in the possibility that, in the worst-case scenario, the agent can learn a greedy solution, hindering any performance gain compared to heuristics such as GreedyGE.

However, since a model trained with the informed reward is able to correctly solve the environment (i.e., finding a path to the identity matrix), we combined the informed reward approach with an additional training phase with the simple reward described in Eq. \ref{eq:reward_simple}, encouraging the agent to choose the right synthesis that minimizes the total number of gates, without focusing exclusively on reducing the overall Hamming distance of the intermediate states. 
From now on, we denote this combination of informed and non-informed reward as \emph{mixed reward}. This was proven to be fundamental for achieving better performance compared to the informed reward, as illustrated by the results of Table \ref{tab:Tab_results_alpha}.
\section{Experimental Results}\label{Section_EXP}

In our experiments we use the same setting for both the only-informed and mixed-reward cases. In particular, both the policy and value networks consist of nine layers with 256 hidden units, for a total of up to 1.1M parameters. The choice of this architecture is motivated by the ablation study, presented in Section \ref{SubSection_EXP_ablation}.

For the unconstrained problem we use both the \textit{informed} and \textit{mixed} rewards as described in Section \ref{sub:Reward Design}. 
In the mixed setting, we use half of the steps for the informed reward, and half for the simple, uninformed one.
For the topology-dependent setting, we use only the \textit{mixed} reward.
In particular, we use $80\%$ of the steps for the informed reward, and the remaining $20\%$ for the uninformed reward.

The number of updates $t$, consisting of a change in the environment state after the application of a chosen action, varies between the two settings: in case of all-to-all connectivity we set it to $500$k, whereas under topological constraints we use $750n$k, where $n$ is the number of qubits (i.e., the matrix size). Note that, as shown in Figure \ref{fig:training_6T}, in most cases a much smaller number of steps is required to correctly solve the environment (i.e., the model is able to synthesize the target matrix).

For both cases, each matrix is generated on-the-fly by starting from the identity and randomly choosing $n^2$ ``moves'' (i.e., the application of a \CNOT). As each matrix can be obtained (in principle) with at most $O\left(\frac{n^2}{\log_2 (n)}\right)$ \CNOTs \citep{pmh2003}, our process ensures a fair sampling over the space of possible matrices. 

For the unconstrained case, the action space consists of each ordered pair of qubits (corresponding to each possible \CNOT$(i,j)$). In the constrained case, the agent is restricted to use \CNOTs according to the considered topology.

Finally, the observation space consists of the set of Boolean matrices of size $n\times n$, corresponding to the space of possible parity matrices as described in Section \ref{sub3:LRC}.

All experiments were conducted on a workstation equipped with an NVIDIA GeForce RTX 3090 GPU (24 GB VRAM) and an Intel Xeon W-2123 CPU (4 cores, 8 threads @ 3.60 GHz) with AVX-512 support and 32 GB of DDR4 RAM. We used Python 3.11.14 and JAX \citep{jax2018github} (jaxlib v0.8.1) for both training and validation. More in detail, we make use of the \texttt{mctx} library \citep{deepmind2020jax} for the JAX-native implementation of the Monte Carlo Tree Search. 

As a starting point, in Section \ref{SubSection_EXP_ablation} we provide a preliminary ablation study on how the quality of synthesized circuits depends on model capacity and the search budget.

We then evaluate our framework on different synthesis tasks. In particular, in Section~\ref{SubSection_EXP_all_to_all} we consider unconstrained linear reversible synthesis (Problem~\ref{Prob1}), while Section~\ref{SubSection_EXP_top} is devoted to topology-aware synthesis (Problem~\ref{Prob2}). In Section~\ref{SubSection_EXP_permutation} we relax the termination criterion by accepting any permutation matrix. Finally, in Section~\ref{SubSection_EXP_phasepolynomials} we move to \CNOT circuits coming from actual applications, namely the trailing circuits obtained by phase-polynomial synthesis of MaxCut and LABS instances.

% ----------------------------------------------------
\subsection{Ablation Analysis}\label{SubSection_EXP_ablation}
In this Section, we investigate how the model's performance scales with respect to the MCTS simulation mechanism, the architectural complexity of the policy and value networks, and the reward function design. Overall, these experiments help us to identify the most effective architectural configuration and quantify the contribution of MCTS in the proposed pipeline. In the fixed setting of unconstrained synthesis with $n=5$ qubits, we vary the number of hidden units per layers in $\{64, 128, 256, 512\}$, and the number of MCTS simulations in $\{1, 10, 50, 100\}$. We report our findings in Table \ref{tab:ablation_n5}.
Notably, \texttt{AlphaCNOT} models where Monte Carlo Tree Search is deactivated, i.e. the number of simulations is set to 1, perform worse than all the others, highlighting the importance of MCTS to achieve better syntheses.
Although models with 512 hidden neurons achieve the best overall results, the benefits are limited compared with the increase in the number of parameters and computational cost. We therefore choose the configuration with 256 hidden neurons and 50 MCTS simulations, which guarantees a reasonable trade-off between solution quality and training cost.

Ultimately, the ablation results showcase the advantage of using the mixed reward over the only-informed one. This also holds for the topology-aware models: after the reward switch, set at 80\% of the training, the mean synthesis length decreases by 9\%-15\% on average.
Additional information on constrained models' performances is available in Figure \ref{fig:training_3_models} in Appendix \ref{Appx:training_process}, where we report 6-Y, 7-L, and 8-T1 training details.

% --------------------------------------------
% --------- Ablation Table -------------------
% --------------------------------------------

\begin{table*}[!t]
    \centering
    \renewcommand{\arraystretch}{.85}

    \begin{tabular}{
    c c
    @{\hspace{8pt}}
    c c
    @{\hspace{14pt}}
    c
}
    \toprule

    & &
    \multicolumn{1}{c}{Informed (inf.)} &
    \multicolumn{1}{c}{Mixed (mix.)} &
    \\

    \cmidrule(lr){3-3}
    \cmidrule(lr){4-4}

    Hidden
    & \#Sim
    & \shortstack{Mean\\CNOT}
    & \shortstack{Mean\\CNOT}
    & \shortstack{Avg. Train\\time (h)} \\
    \midrule

    \multirow{4}{*}{64}
    & 1   & 10.40 & 10.40 & 0h 06m \\
    & 10  & 8.71  & 8.55  & 0h 18m \\
    & 50  & 8.65  & 8.40  & 1h 51m \\
    & 100 & 8.55  & 8.25  & 6h 41m \\

    \addlinespace

    \multirow{4}{*}{128}
    & 1   & 9.97 & 9.96 & 0h 07m \\
    & 10  & 8.52 & 8.30 & 0h 19m \\
    & 50  & 8.51 & 8.21 & 1h 55m \\
    & 100 & 8.43 & 8.10 & 7h 35m \\

    \addlinespace

    \multirow{4}{*}{\textbf{256}}
    & 1   & 9.63 & 9.59 & 0h 08m \\
    & 10  & 8.37 & 8.17 & 0h 21m \\
    & \textbf{50}
    & \textbf{8.35}
    & \textbf{8.11}
    & \textbf{2h 17m} \\
    & 100 & 8.32 & 8.05 & 7h 51m \\

    \addlinespace

    \multirow{4}{*}{512}
    & 1   & 9.73 & 9.61 & 0h 10m \\
    & 10  & 8.32 & 8.11 & 0h 28m \\
    & 50  & 8.31 & 8.08 & 2h 32m \\
    & 100 & 8.28 & 8.02 & 9h 14m \\

    \bottomrule
\end{tabular}

    \caption{Ablation study for the unconstrained $n=5$ case.
        Mean \CNOT count are measured at the end of training over the last 5\% of the training budget. We average the two training times, measured on the same hardware, since no significant delay is due to the reward switch. We highlight in bold the parameters choice adopted as standard for all \texttt{AlphaCNOT} models.
        }
    \label{tab:ablation_n5}
\end{table*}

% -----------------------------------------------------
% ALL-TO-ALL
\subsection{Unconstrained \CNOT Synthesis}\label{SubSection_EXP_all_to_all}
To assess our method, we test the agent for $n\in\{4,5,6,7,8\}$ qubits. We compare the average number of \CNOTs obtained by \texttt{AlphaCNOT} with the results from PMH \citep{pmh2003} (implemented on \emph{Qiskit} \citep{qiskit2024}), AECM \citep{schaeffer2014}, GreedyGE \citep{debrugiere2021}, RowCol \citep{wu2023rowcol}, and the PPO-based  Gaussian Striping method proposed in \cite{romanello2025}, denoted as RL-GS, where the agent is evaluated $100$ times for each problem instance to select the best result.
We benchmark our dimension-based RL agents, denoted as ``\texttt{AlphaCNOT} (inf.)'' and ``\texttt{AlphaCNOT} (mix.)'', against state-of-the-art heuristic methods. 

As RL is inherently non-deterministic, for each instance we evaluate the model $100$ times (denoted as ``\texttt{AlphaCNOT}$_{100}$''), similarly to what is done in \cite{romanello2025} and \cite{kremer2024}. For completeness, we report also the results for the model sampled once. 
Finally, for smaller values of $n$ we present the optimal solution, found by the ASP model of \cite{piazza2023}. Note that computing the optimal values for higher $n$ was unfeasible under a reasonable timeout threshold of one hour per circuit.

The complete performance metrics are reported in Table \ref{tab:Tab_results_alpha}.

\begin{table*}[htb]
    \centering
    \begin{tabular}{lccccc}
        \toprule
        & \multicolumn{5}{c}{Problem Size ($n$)} \\ 
        \cmidrule(lr){2-6}
        Approach & 4 & 5 & 6 & 7 & 8 \\ 
        \midrule
        Optimal (ASP)  & 5.28 & 8.01 & 10.64 & - & - \\
        \addlinespace
        PMH  & 6.98 & 11.07 & 16.40 & 22.62 & 30.58  \\ 
        AECM  & 6.61 & 10.58 & 15.34 & 21.08 &27.51  \\ 
        GreedyGE  & 7.94 & 12.34 & 17.49 & 23.40 & 29.81  \\ 
        RowCol  & 7.30 & 11.94 & 17.71 & 24.41 & 31.38\\
        RL-GS$_{100}^*$  & 7.19 & 11.84 & 16.20 & 22.61 & 28.02  \\
        \texttt{AlphaCNOT}$_1$ (inf.)  & 5.32 & 8.42 & 12.07 & 16.79 & 23.28 \\
        \texttt{AlphaCNOT}$_1$ (mix.)  & \textbf{5.29} & \textbf{8.16} & \textbf{11.50} & \textbf{16.15} & \textbf{23.05} \\
        
        \addlinespace
        
        \texttt{AlphaCNOT}$_{100}$ (inf.)  & 5.32 & 8.26 & 11.38 & 15.49 & \textbf{20.11} \\
        \texttt{AlphaCNOT}$_{100}$ (mix.)  & \textbf{5.28} & \textbf{8.06} & \textbf{10.97} & \textbf{14.99} & 20.62 \\

        \bottomrule
    \end{tabular}
    \caption{Average \CNOT count over the same set of 100 random instances. Optimal values are computed with ASP \citep{piazza2023}, where feasible. PMH results are obtained from the Qiskit library \citep{qiskit2024}.  AECM, GreedyGE, and RowCol are obtained by custom implementation of the algorithms \citep{schaeffer2014, debrugiere2021, wu2023rowcol} . The  RL-GS results, denoted with $^*$, are reported from the corresponding paper \citep{romanello2025} and correspond to the average over different random instances. Best results both in the 1-shot and 100-shot fashion are highlighted in bold.}
    \label{tab:Tab_results_alpha}
\end{table*}

We can observe that our RL approach efficiently addresses the task, as even when used in a one-shot fashion, our method performs better than all the other algorithms. 
Finally, we observe that the advantage of \texttt{AlphaCNOT} (1-shot, mix.) compared to PMH remains consistent through the problem dimension scale, with a reduction in \CNOT gate count ranging from $24.21\%$ for $n=4$ to $24.62\%$ for $n=8$.
In the 1-shot setting, the mixed approach performs better than the informed approach. In the 100-shot setting, the mixed approach achieves superior performance in all cases except for the $n=8$ case, where the two approaches exhibit comparable performance.

\subsection{Topology Constraints}\label{SubSection_EXP_top}
\begin{figure}[htbp]
    \centering
    \includegraphics[width=0.60\linewidth]{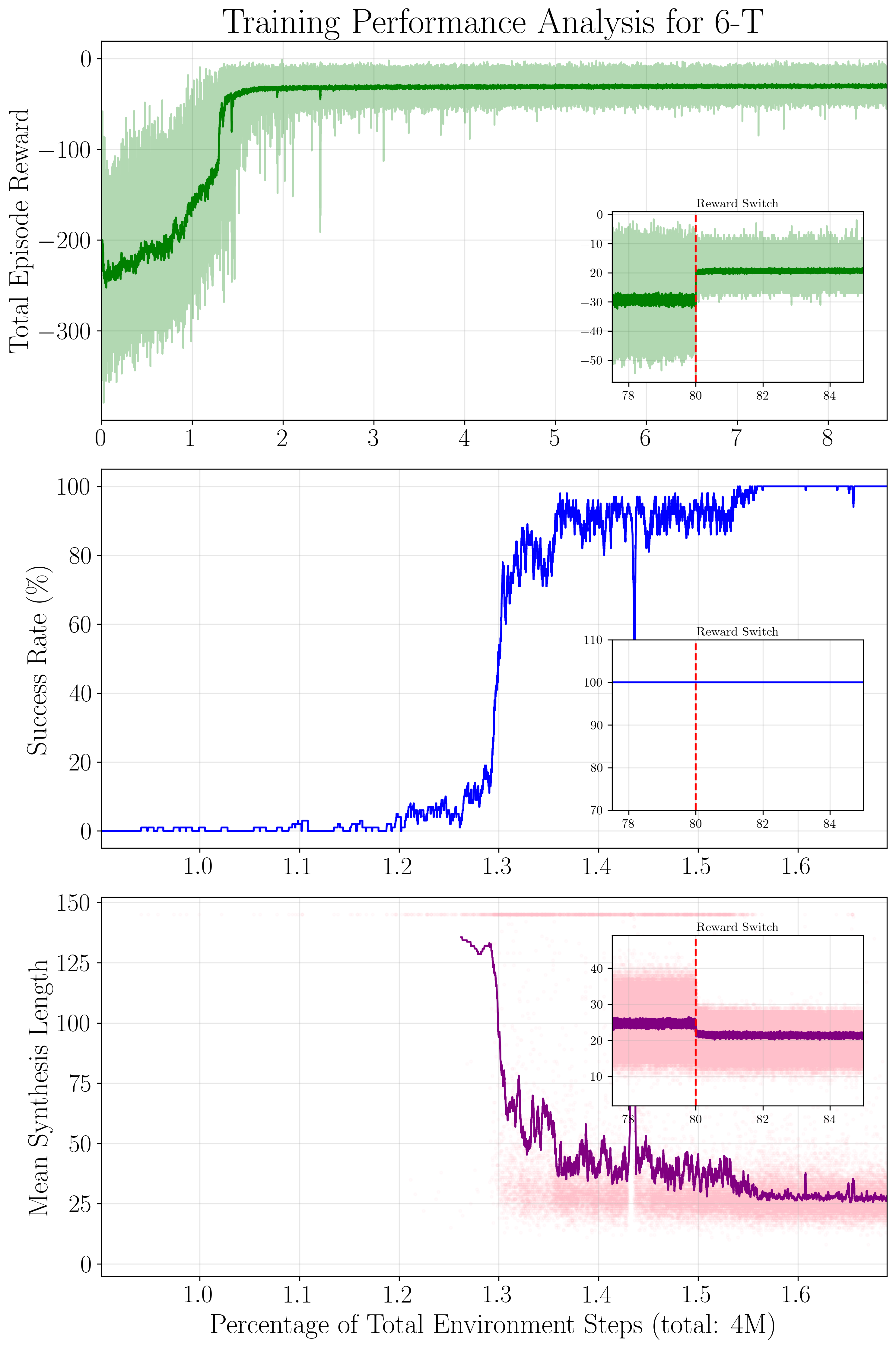}
    \caption{Training Dynamics of ``6-T'' model. The green upper graph exhibits the total episode reward, while the blue central and purple lower curves describe the success rate and mean synthesis length, respectively. The reward switch, set at 80\% of the total environment steps, is highlighted by a red dashed line in the inset plots beside each panel.}
    \label{fig:training_6T}
\end{figure}

To illustrate the learning dynamics of our approach, we first analyze the training metrics for a representative topology-constrained instance. Figure \ref{fig:training_6T} shows, from top to bottom, the total episode reward, the success rate, and the mean synthesis length for the ``6-T'' case. 

We now evaluate our method in the topology-constrained case, following the experimental setting provided in \cite{kremer2024}.
In particular, we evaluate \texttt{AlphaCNOT} on selected configurations of realistic hardware topologies for up to $8$ qubits, derived from the following templates: Linear (L), Y-form, T-form, H-form, and F-form (see Figure \ref{fig:appendix_topologies} in Appendix \ref{Appx:topology} for further details). 

We compare our results with those based on curriculum learning provided in \cite{kremer2024}, denoted as RL-CL, though calculated on different instances. In addition, we evaluate as a non-RL solution PMH combined with SABRE \citep{zou2024} (where the PMH approach is first applied to the unconstrained setting and SABRE routing is performed afterwards), and with the optimal solution found with ASP on up to $6$ qubits.
The resulting performance metrics are summarized in Table~\ref{Tab_results_alpha_TOP}.

Lastly, we provide the validation time of \texttt{AlphaCNOT} and RL-CL \citep{kremer2024} in Appendix \ref{Appx:time_analysis}, Figure \ref{fig:time_graphics_top}.

\begin{table*}[t]
    \centering
    \small
    \setlength{\tabcolsep}{4pt}

    % First half
    \begin{adjustbox}{max width=\textwidth}
    \begin{tabular}{l *{7}{c}}
        \toprule
        Method
        & 4-L & 4-Y & 5-L & 5-T & 6-L & 6-T & 6-Y \\
        \midrule

        Optimal 
        & 8.96 & 7.37 & 15.18 & 13.00 & 23.33 & 20.50 & 19.76 \\

        PMH+SABRE
        & 15.6 & 12.9 & 29.9 & 24.8 & 53.3 & 45.8 & 44.4 \\

        Steiner
        & 10.93 & 10.22 & 19.61 & 18.49 & 31.23 & 29.28 & 30.45 \\
        
        RowCol 
        & 11.21 & 10.02 & 20.03 & 18.68 & 31.29 & 29.78 & 28.36 \\

        \addlinespace

        RL-CL$_1$ 
        & 10.2 & 8.3 & 17.2 & 14.8 & 27.1 & 23.9 & 23.1 \\

        \texttt{AlphaCNOT}$_1$ (mix.)
        & \textbf{8.97} & \textbf{7.37} & \textbf{15.46} & \textbf{13.23}
        & \textbf{24.54} & \textbf{21.47} & \textbf{20.95} \\

        \addlinespace

        RL-CL$_{100}$
        & 10.0 & 8.1 & 16.1 & 13.9 & 25.4 & 22.5 & 21.6 \\

        \texttt{AlphaCNOT}$_{100}$ (mix.)
        & \textbf{8.97} & \textbf{7.37} & \textbf{15.24} & \textbf{13.03}
        & \textbf{23.44} & \textbf{20.66} & \textbf{19.89} \\

        \bottomrule
    \end{tabular}
    \end{adjustbox}

    \vspace{0.6em}

    % Second half
    \begin{adjustbox}{max width=\textwidth}
    \begin{tabular}{l *{7}{c}}
        \toprule
        Method
        & 7-L & 7-T & 7-Y & 8-H & 8-F & 8-T1 & 8-T2 \\
        \midrule

        PMH+SABRE
        & 84.3 & 76.2 & 67.9 & 104.2 & 116.3 & 123.5 & 106.3 \\

        Steiner
        & 45.95 & 43.84 & 45.44 & 64.46 & 63.54 & 59.77 & 64.15 \\

        RowCol 
        & 45.50 & 43.90 & 42.41 & 54.36 & 60.35 & 61.53 & 57.14 \\

        \addlinespace
        
        RL-CL$_1$
        & 40.1 & 36.7 & 34.4 & 48.9 & 52.2 & 54.1 & 50.6 \\

        \texttt{AlphaCNOT}$_1$ (mix.)
        & \textbf{37.48} & \textbf{33.36} & \textbf{31.54}
        & \textbf{42.40} & \textbf{46.35} & \textbf{49.18} & \textbf{43.21} \\

        \addlinespace

        RL-CL$_{100}$ 
        & 37.5 & 34.3 & 31.0 & 45.0 & 47.6 & 49.5 & 45.4 \\

        \texttt{AlphaCNOT}$_{100}$ (mix.)
        & \textbf{34.67} & \textbf{31.01} & \textbf{28.55}
        & \textbf{38.70} & \textbf{42.03} & \textbf{44.82} & \textbf{39.19} \\

        \bottomrule
    \end{tabular}
    \end{adjustbox}

    \caption{Average \CNOT count over 100 instances under topology constraints.
    Optimal \CNOT counts up to 6 qubits are computed with ASP \citep{piazza2023top}. Steiner results are obtained with the topology-aware \textsc{Steiner-Gauss} method of \cite{kissinger2019}, whereas RowCol results are obtained with the algorithm of \cite{wu2023rowcol}. PMH results are implemented using the Qiskit library by first applying
    PMH for unconstrained decomposition and then SABRE for routing, consistently with \cite{kremer2024}. RL-CL (Reinforcement Learning with Curriculum Learning) results are taken from \cite{kremer2024} and were not recomputed on
    the same 100 instances used to evaluate the other methods. \texttt{AlphaCNOT} performance data, both 1-shot and 100-shot, refer solely to the mixed-reward variant.
    }
    \label{Tab_results_alpha_TOP}
\end{table*}

\subsection{\texttt{PermAlphaCNOT}}\label{SubSection_EXP_permutation}
Up to this point, we require CNOT synthesis of the parity matrix to reach the identity state specifically. Following PermRowCol \citep{meijer2023permrowcol}, we can relax the terminal condition of \texttt{AlphaCNOT} by accepting any permutation matrix.
Essentially,
our objective is to find a \CNOT circuit with parity matrix $M_\mathcal{C}$
such that $M_\mathcal{C} = M_{\mathcal{C}_{\mathrm{target}}}\cdot P$, where $P$ is a
permutation matrix. Such a matrix contains exactly one 1 per column, thus it satisfies the following condition
\begin{equation*}
    \forall{i}\in\{1,\ldots,n\} \,\,\sum_{j=1}^{n} P_{ij}=1
\end{equation*}
or, equivalently, there are exactly $n$ non-zero entries in the matrix.
All models are trained in a mixed fashion, with the reward switch set at $50\%$ and $80\%$ of the training budget for the unconstrained and topology-aware settings, respectively. The non-informed reward penalizes every non-terminal state with $-1$ and rewards the agent with $+1$ when a terminal state is reached.

From now on, we will refer to this generalization of \texttt{AlphaCNOT} as \texttt{PermAlphaCNOT}. The latter variant is particularly relevant under topological constraints, where restoring the original qubit ordering may require many routing operations. However, it is also meaningful in the all-to-all setting: although routing is unnecessary, the permutation-relaxed setting was shown to produce shorter \CNOT decompositions. The results for unconstrained and topology-aware synthesis are listed in Tables \ref{tab:results_alpha_permutation_all_to_all} and \ref{tab:results_alpha_permutation_topology}, respectively.

\begin{table*}[htb]
\centering
\begin{tabular}{lccccc}
\toprule
& \multicolumn{5}{c}{Problem Size ($n$)} \\
\cmidrule(lr){2-6}
Approach & 4 & 5 & 6 & 7 & 8 \\
\midrule
PermRowCol  & 4.52 & 7.66 & 11.82 & 17.36 & 23.20 \\
PermRowCol-RT$_{15}$  & 4.37 & 7.45 & 11.69 & 17.30 & 23.13 \\
\addlinespace
\texttt{PermAlphaCNOT}$_1$ (mix.) & \textbf{3.43} & 5.41 & 8.19 & 15.41 & 22.20 \\
\texttt{PermAlphaCNOT}$_{100}$ (mix.) & \textbf{3.43} & \textbf{5.38} & \textbf{7.99} & \textbf{14.08} & \textbf{19.61} \\
\bottomrule
\end{tabular}
\caption{Average CNOT count for permutation-invariant unconstrained synthesis over the same 100 matrices. PermAlphaCNOT$_{1}$ with one shot reports one synthesis per matrix; PermAlphaCNOT$_{100}$ reports the best of 100 syntheses per matrix. PermRowCol is deterministic and is therefore evaluated once per matrix. PermRowCol-RT (with Reverse Transversal strategy) is a variant of PermRowCol proposed in \cite{meijer2023permrowcol}. We use 15 RT iterations, since no better syntheses were found with higher values.}
\label{tab:results_alpha_permutation_all_to_all}
\end{table*}

\begin{table}[h!]
    \centering
    \small
    \setlength{\tabcolsep}{4pt}

    % First half
    \begin{adjustbox}{max width=\textwidth}
    \begin{tabular}{l *{7}{c}}
        \toprule
        Method
        & 4-L & 4-Y & 5-L & 5-T & 6-L & 6-T & 6-Y \\
        \midrule

        PermRowCol 
        & 7.78 & 6.41 & 15.06 & 13.38 & 25.64 & 23.99 & 22.36 \\

        PermRowCol-RT$_{15}$ 
        & 7.64 & 6.33 & 14.64 & 13.09 & 25.32 & 23.47 & 21.99 \\

        \addlinespace

        \texttt{PermAlphaCNOT}$_1$ (mix.)
        & \textbf{6.21} & \textbf{5.18} & \textbf{10.94}
        & 9.40 & 17.69 & 15.59 & 15.04 \\

        \texttt{PermAlphaCNOT}$_{100}$ (mix.)
        & \textbf{6.21} & \textbf{5.18} & \textbf{10.94}
        & \textbf{9.39} & \textbf{17.36} & \textbf{15.36} & \textbf{14.73} \\

        \bottomrule
    \end{tabular}
    \end{adjustbox}

    \vspace{0.6em}

    % Second half
    \begin{adjustbox}{max width=\textwidth}
    \begin{tabular}{l *{7}{c}}
        \toprule
        Method
        & 7-L & 7-T & 7-Y & 8-H & 8-F & 8-T1 & 8-T2 \\
        \midrule

        PermRowCol 
        & 38.82 & 36.93 & 34.62 & 49.26 & 51.87 & 55.07 & 51.26 \\
        
        PermRowCol-RT$_{15}$ 
        & 38.76 & 36.80 & 34.48 & 49.08 & 51.68 & 54.95 & 51.19 \\

        \addlinespace

        \texttt{PermAlphaCNOT}$_1$ (mix.)
        & 26.98 & 24.34 & 22.19
        & 32.13 & 35.25 & 38.12 & 33.14 \\

        \texttt{PermAlphaCNOT}$_{100}$ (mix.)
        & \textbf{25.90} & \textbf{23.41} & \textbf{21.04}
        & \textbf{30.13} & \textbf{32.75} & \textbf{35.00} & \textbf{30.71} \\

        \bottomrule
    \end{tabular}
    \end{adjustbox}

    \caption{Average CNOT count over the same 100 instances. All methods synthesize each target matrix up to a permutation matrix, while respecting the corresponding connectivity graph. PermRowCol~\citep{meijer2023permrowcol} is deterministic and is therefore evaluated once per instance. PermRowCol-RT$_{15}$ applies the Reverse Traversal strategy of~\citep{meijer2023permrowcol} for 15 iterations. Both \texttt{PermAlphaCNOT} results refer to the mixed-reward variant. Best results for each topology are highlighted in bold.
    }
    \label{tab:results_alpha_permutation_topology}
\end{table}

\subsection{Synthesis of Trailing \CNOT Circuits from QAOA}
\label{SubSection_EXP_phasepolynomials}

We now consider as application case the synthesis of trailing \CNOT circuits obtained from Quantum Approximate Optimization Algorithm (QAOA)~\citep{QAOA_2014} circuits. In particular, we will focus on two different problem classes: MaxCut \citep{MaxCut_1995} and Low Autocorrelation Binary Sequence (LABS) \citep{LABS_1987}. For MaxCut, we generate randomly $10$ instances of $3-$regular graphs for each problem size $n\in\{4, 6, 8\}$, corresponding to circuits with $n$ qubits. For LABS, we consider the unique instance of size $n$ for each topology with $n \in \{4, 5, 6, 7, 8\}$. Additional details on the consider problem classes and how they are formalized in QAOA are provided in Appendix \ref{Appx:qaoa}.

Then, for each instance, we first build the corresponding QAOA circuit layer. Then, we compile the resulting layer with the phase-polynomial synthesis algorithm {ParitySynth}~\citep{vandaele2022phasesynthesis}, as implemented in \textit{pauliopt}~\citep{gogioso2021pauliopt}. Finally, we consider the resulting trailing \CNOT circuit and we optimize it with \texttt{AlphaCNOT}.

\begin{figure}[h!]
    \centering
    \includegraphics[width=\linewidth]
        {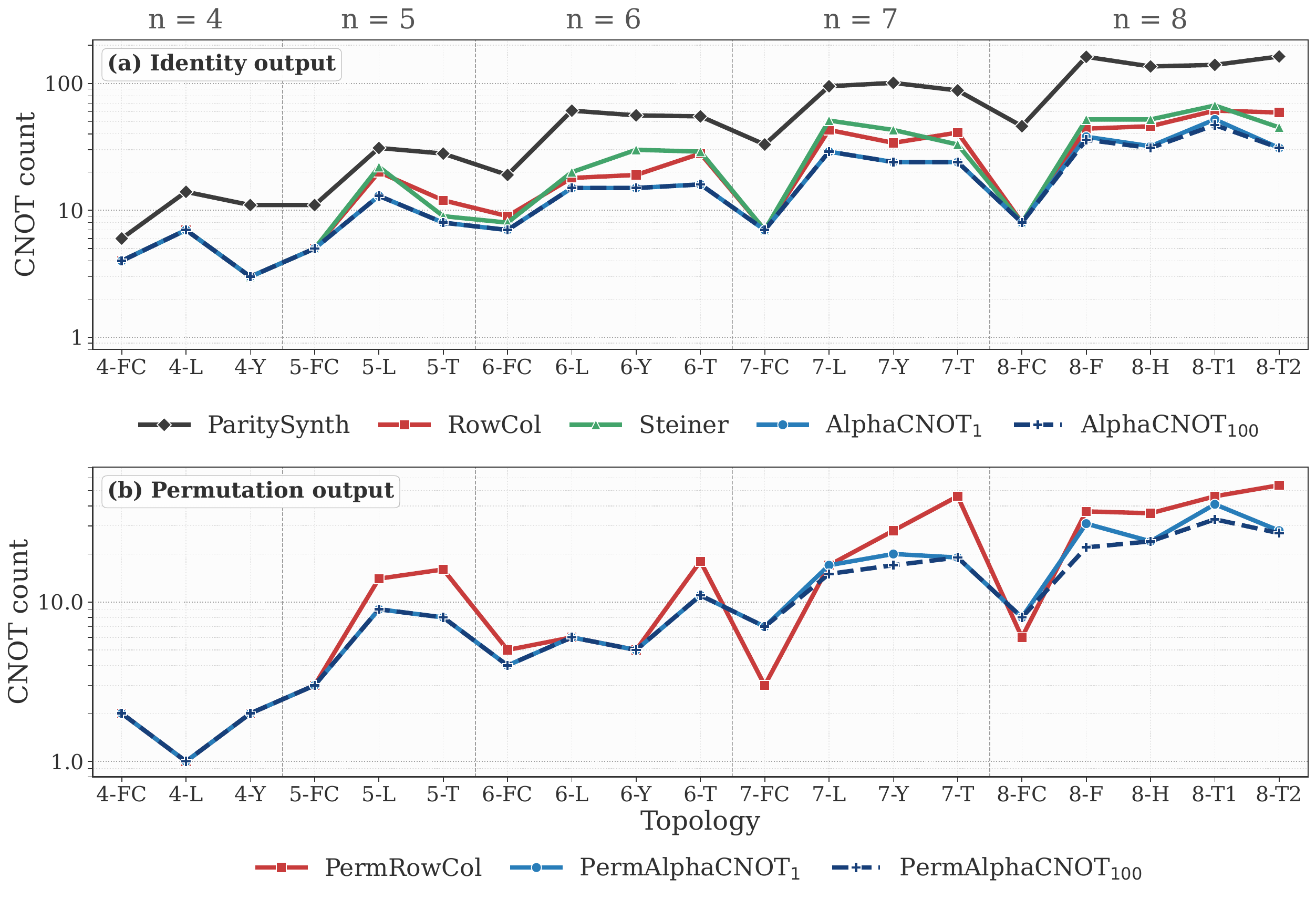}

    \caption{Results for the LABS instances across 19 topologies.
    Panel (a) compares the original ParitySynth \CNOT tail,
    \texttt{AlphaCNOT}$_1$, \texttt{AlphaCNOT}$_{100}$, and PermRowCol
    with restoration in the identity setting.
    Panel (b) compares \texttt{PermAlphaCNOT}$_1$,
    \texttt{PermAlphaCNOT}$_{100}$, and PermRowCol in the
    permutation-relaxed setting. PermRowCol (+restoration)
    is obtained via \textit{pauliopt}~\citep{gogioso2021pauliopt}.}
    \label{fig:results_LABS}
\end{figure}

We evaluate our method on all the topologies considered in this work (including the fully connected ones), and for both the standard and permutation-free settings.
We compare our results with two synthesis methods available in \textit{pauliopt}, namely RowCol~\citep{wu2023rowcol}, and PermRowCol~\citep{meijer2023permrowcol}, and Steiner from \cite{kissinger2019}. 
Results in Table \ref{tab:results_MaxCut} evidence a stable advantage of \texttt{AlphaCNOT} and \texttt{PermAlphaCNOT} across all topologies except for the unconstrained $n=8$ case, in which PermRowCol performs slightly better than our model in the 100-shot setting. Lastly, in Figure~\ref{fig:results_LABS}, we show that
\texttt{PermAlphaCNOT} generally achieves significantly lower \CNOT counts
than PermRowCol. However, for the fully connected topologies
with $n=7$ and $n=8$, \texttt{PermAlphaCNOT} struggles to identify
efficient synthesis strategies for matrices that are close to
the identity. We can attribute this behaviour to the training distribution, since our RL model is primarily trained on complex dense matrices.

% MaxCut
\begin{table}[htbp]
    \centering
    \small
    \setlength{\tabcolsep}{4pt}
    \begin{adjustbox}{max width=\textwidth}
        \begin{tabular}{lc cccc ccc}
        \toprule
        & & \multicolumn{4}{c}{Identity output}
        & \multicolumn{3}{c}{Permutation output} \\
        \cmidrule(lr){3-6}\cmidrule(lr){7-9}
        Topology
        & \shortstack{ParitySynth}
        & RowCol
        & Steiner
        & \shortstack{\texttt{AlphaCNOT}$_1$}
        & \shortstack{\texttt{AlphaCNOT}$_{100}$}
        & PermRowCol
        & \shortstack{\texttt{PermAlphaCNOT}$_1$}
        & \shortstack{\texttt{PermAlphaCNOT}$_{100}$} \\
        \midrule

        4-FC &  9.0 & \textbf{4.0} & \textbf{4.0} & \textbf{4.0} & \textbf{4.0} & 4.0 & \textbf{3.0} & \textbf{3.0} \\
        4-L  & 15.0 & 15.0 & 13.0 & \textbf{9.0} & \textbf{9.0} & 11.0 & \textbf{7.0} & \textbf{7.0} \\
        4-Y  &  9.0 &  8.0 &  7.0 & \textbf{5.0} & \textbf{5.0} & \textbf{4.0} & \textbf{4.0} & \textbf{4.0} \\

        \addlinespace

        6-FC & 17.5 &  7.1 &  7.5 & \textbf{6.9} & \textbf{6.9} & 6.9 & \textbf{4.8} & \textbf{4.8} \\
        6-L  & 41.8 & 24.3 & 25.7 & 18.3 & \textbf{18.1} & 21.2 & \textbf{12.5} & \textbf{12.5} \\
        6-T  & 41.9 & 20.5 & 21.6 & 16.6 & \textbf{16.4} & 15.4 & 11.0 & \textbf{10.7} \\
        6-Y  & 37.3 & 24.8 & 25.8 & 17.9 & \textbf{17.3} & 14.5 & \textbf{11.3} & \textbf{11.3} \\

        \addlinespace

        8-FC & 25.7 & 13.1 & \textbf{12.5} & 12.9 & 12.7 & \textbf{11.0} & 12.9 & 11.9 \\
        8-H  & 61.6 & 41.2 & 49.1 & 26.9 & \textbf{25.9} & 29.4 & 19.5 & \textbf{19.1} \\
        8-F  & 68.0 & 44.3 & 48.8 & 36.0 & \textbf{33.6} & 37.4 & 25.6 & \textbf{24.1} \\
        8-T1 & 73.2 & 38.4 & 40.9 & 28.9 & \textbf{27.6} & 30.3 & 21.9 & \textbf{20.5} \\
        8-T2 & 65.7 & 36.7 & 43.3 & 27.1 & \textbf{25.6} & 32.1 & 20.7 & \textbf{19.4} \\

        \bottomrule
    \end{tabular}
    \end{adjustbox}

    \caption{Average \CNOT count of the trailing circuit for the
    MaxCut cases over 10 instances. ``FC'' topology stands for fully connected. Our models are trained with the mixed reward,
    The ParitySynth column reports the original \CNOT tail average before
    re-synthesis. We highlight the best results in bold, both for identity and permutation settings.}
    \label{tab:results_MaxCut}
\end{table}

\section{Conclusion}\label{Section_CONCLUSION}

As we steadily progress toward the so-called ``quantum utility'' \citep{eisert2025,herrmann2023} phase of quantum computing, in which quantum technology will become a practical and reliable tool for research and industry, quantum resource optimization will be a central issue to address.
In this regard, we addressed the \CNOT minimization problem as a prime example of circuit optimization.

By combining reinforcement learning with MCTS strategies, our model \texttt{AlphaCNOT} obtained a substantial reduction in terms of \CNOT gates per circuit, compared to both heuristic methods (such as PMH \citep{pmh2003}, AECM \citep{schaeffer2014}, and GreedyGE \citep{debrugiere2021}) and other PPO-based approaches (like \cite{romanello2025} and \cite{kremer2024}). These promising results encourage the use of model-based RL algorithms when dealing with challenging problems with high-dimensional search spaces, in which the complete exploration is computationally infeasible. 

It is interesting to note that, despite our model being precisely designed for finding syntheses of circuits composed solely of \CNOT gates, the idea of addressing circuit synthesis with RL models based on AlphaZero \citep{silver2017zero} is quite general,  and could be easily applied to other problems of quantum optimization that are representable as planning tasks, 
including qubit routing and mapping, hardware-aware circuit synthesis, and circuit optimization through \emph{gadgets}, i.e., local rewriting rules.
Lastly, a promising direction for future research is the incorporation of equivariant neural network architectures, following \cite{Richie_2026} ideas, into the AlphaCNOT framework. These models could improve generalization across different circuit sizes and enhance the scalability of learning-based synthesis methods.

In conclusion, our  results suggest that reinforcement learning will become a central instrument for quantum optimization, with circuit synthesis representing one of its main challenges. These improvements could be an initial step in the transition toward quantum utility in the near future, where quantum devices will play a crucial role in addressing classes of problems that exceed the capabilities of classical computers.

\bibliography{bibfile}
\appendix
\clearpage
\begin{appendices}\label{Section_appendix}
\appendixpage

This section provides supplementary materials, metrics and experimental results that support our algorithm, \texttt{AlphaCNOT}. The content is organized as follows: Appendix \ref{Appx:time_analysis} presents a detailed time complexity and performance analysis of the methods employed in the topological setting; Appendix \ref{Appx:training_process} reports additional insights and metrics regarding the training performance of the topological case ``6-Y'',``7-L'', and ``8-T1'';  Appendix \ref{Appx:parameters} contains all the training parameters of the RL models; in Appendix \ref{Appx:qaoa} we provide additional details on QAOA and the problems considered for phase polynomial synthesis; finally, Appendix \ref{Appx:topology} describes the topological structures adopted in our study.

\section{Inference Time Analysis}\label{Appx:time_analysis}
\begin{figure*}[htb]
    \centering
    \includegraphics[width=\linewidth]{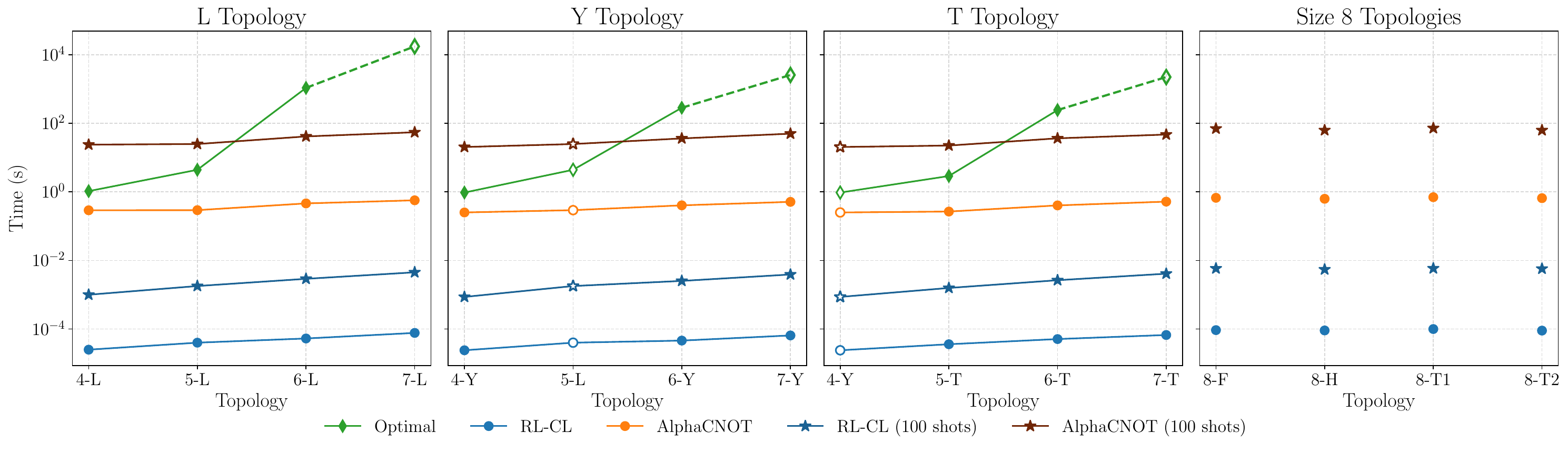}
    \caption{Inference time evaluation across all tested topologies. Results for RL-CL are reported from \cite{kremer2024}. The time for the optimal solution in ASP is computed for topologies up to 6 qubits. For 7 qubits, the ASP runtime is estimated via regression and clearly represents an underestimate. Values with empty marks (e.g. 5-L in the second plot with Y topologies) are selected between topologies with highest similarity (see Figure \ref{fig:appendix_topologies}).}
    \label{fig:time_graphics_top}
\end{figure*}

Figure \ref{fig:time_graphics_top} reports the inference times for  RL-CL (as reported in \cite{kremer2024}) and \texttt{AlphaCNOT} on Problem \ref{Prob2}. We also report the time required to find the optimal solution with ASP (in particular, times for up to 6 qubits are computed on our machine, while the time for 7-qubit instances is estimated via an exponential regression, leading to an underestimation of the real required time).

We note that \texttt{AlphaCNOT} requires more time than what is reported for RL-CL. Despite the difference in hardware playing a relevant role in this discrepancy, it is important to note that tree-based search is, in general, computationally  complex. However, even when used in a 100-shot fashion, our model required less than 2 minutes per circuit, with small differences between different topologies and qubit counts.

On the other hand, optimal solutions for 6 qubits required up to 30 minutes. A conservative estimation suggests that for 7 qubits, ASP would require at least 2 hours per circuit.

\section{Additional Training Process Analysis}\label{Appx:training_process}

\begin{figure*}[htb]
    \centering
    \begin{minipage}{0.32\linewidth}
        \centering
        \includegraphics[width=\linewidth]{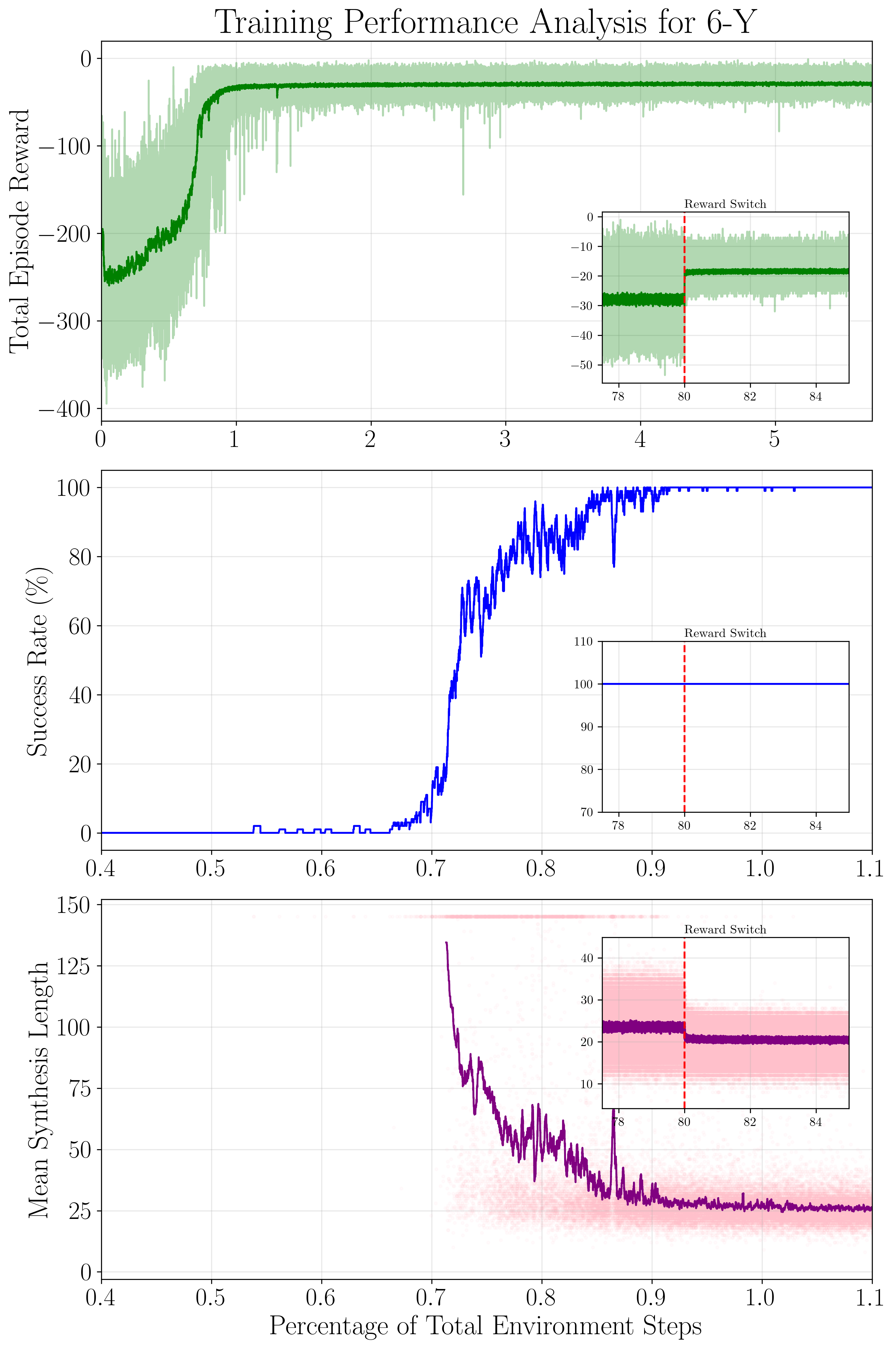}
        \caption*{(a) 6-Y topology}
    \end{minipage}
    \hfill
    \begin{minipage}{0.32\linewidth}
        \centering
        \includegraphics[width=\linewidth]{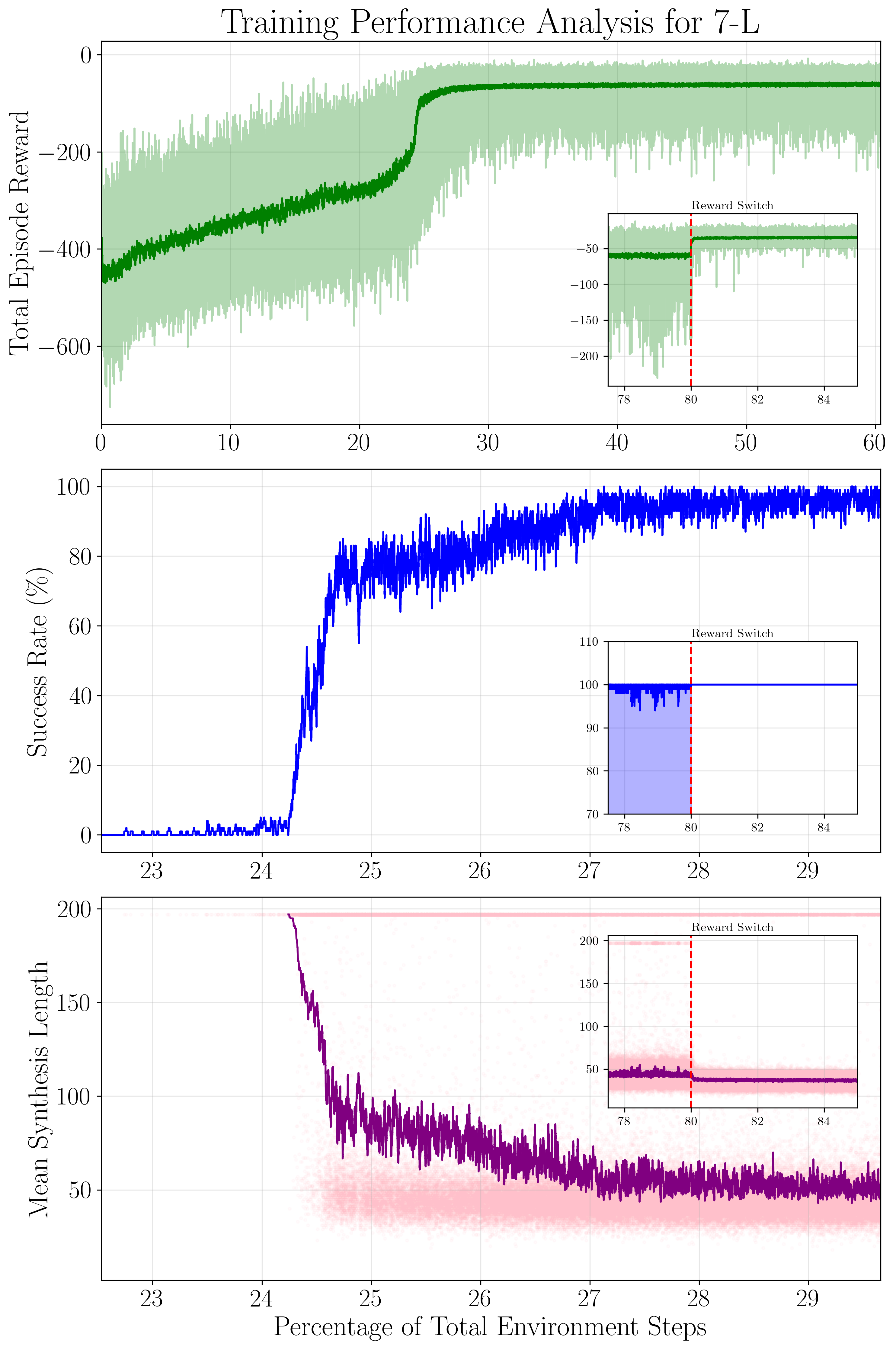}
        \caption*{(b) 7-L topology}
    \end{minipage}
    \hfill
    \begin{minipage}{0.32\linewidth}
        \centering
        \includegraphics[width=\linewidth]{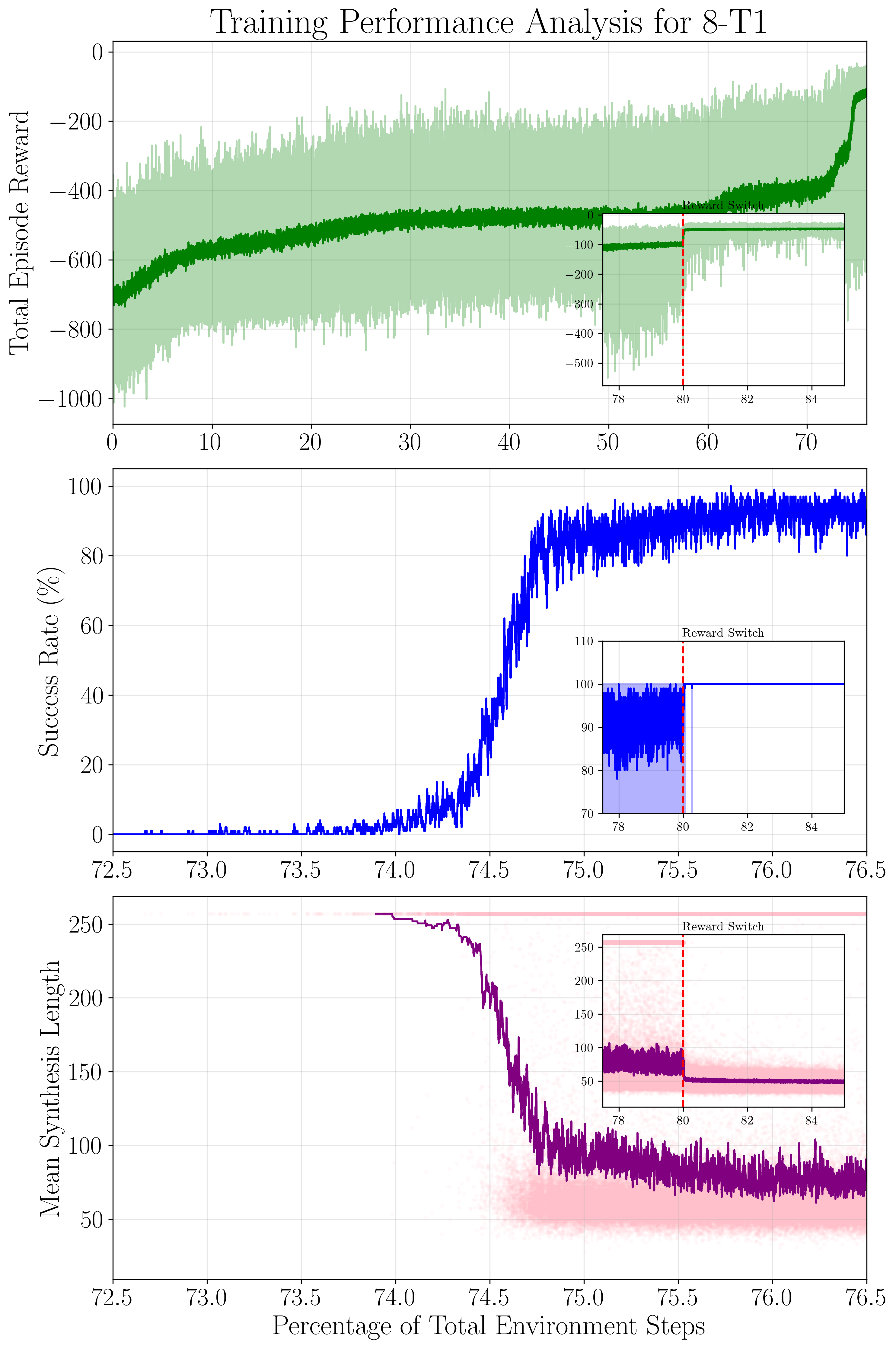}
        \caption*{(c) 8-T1 topology}
    \end{minipage}

    \caption{Training dynamics of ``6-Y'', ``7-L'', and ``8-T1'' models shown as three parallel panels.The green upper graph represents the total episode reward, while the blue central and purple lower curves describe the success rate and mean synthesis length, respectively. The red dashed line highlights the reward switch set at 80\% of total time steps.}
    \label{fig:training_3_models}
\end{figure*}

Figure \ref{fig:training_3_models} reports three panels showing the training metrics of the \texttt{AlphaCNOT} models across the ``6-Y'', ``7-L'', and ``8-T1'' topologies. After the reward switch, set at 80\% of the total time steps, the total episode reward improves without compromising the success rate. In fact, in all cases the model succeeded in reaching a 100\% success rate within the initial informed phase, except for the ``8-T1'' topology, which is particularly challenging. In this scenario, the mixed reward is fundamental in order to achieve a complete understanding of the problem's dynamics, enabling the model to achieve a 100\% success rate and effectively reduce the mean synthesis length by a significant 23\%.

\section{Parameter Analysis}\label{Appx:parameters}
We trained a separate policy-value pair for every problem size, reward design, and hardware topology. In what follows, we report the set of parameters of every \texttt{AlphaCNOT} model we used, including optimization hyperparameters and models' parameters, in Table \ref{tab:alphacnot_hyperparameters} and \ref{tab:alphacnot-model-inventory}, respectively.

\begin{table*}[t]
\centering
\small
\begin{tabularx}{\textwidth}{@{}p{0.29\textwidth}X@{}}

\toprule
Parameter & Value and definition \\
\midrule
Optimizer & AdamW optimizer, $\beta_1=0.9$, $\beta_2=0.99$, $\epsilon=10^{-5}$, weight decay $10^{-6}$. \\
Learning rates & $4\times10^{-5}$ (policy) and $10^{-4}$ (value) \\
Batch size & 64 \\
Discount factor & $\gamma=0.99$\\
MCTS simulations & 50 (training and evaluation), $\{1,10,50,100\}$ (ablation)\\
\bottomrule

\end{tabularx}

\caption{Optimization hyperparameters used by \texttt{AlphaCNOT}.}
\label{tab:alphacnot_hyperparameters}
\end{table*}

\begin{table*}[t]
\centering
\scriptsize
\setlength{\tabcolsep}{3.2pt}
\renewcommand{\arraystretch}{1.30}

\begin{tabular}{@{}cc ccccc@{}}
\hline
\rule{0pt}{2.8ex}
Setting
& Models
& $d_{\rm in}$
& $|\mathcal{A}|$
& $|\theta_\pi|$
& $|\theta_V|$
& $|\theta|$ \\
\hline

\multirow{5}{*}{Fully connected}
& \rule{0pt}{2.7ex}$n=4$
& 16 & 12 & 533{,}772 & 530{,}945 & 1{,}064{,}717 \\
& $n=5$ & 25 & 20 & 538{,}132 & 533{,}249 & 1{,}071{,}381 \\
& $n=6$ & 36 & 30 & 543{,}518 & 536{,}065 & 1{,}079{,}583 \\
& $n=7$ & 49 & 42 & 549{,}930 & 539{,}393 & 1{,}089{,}323 \\
& $n=8$ & 64 & 56 & 557{,}368 & 543{,}233 & 1{,}100{,}601 \\[2pt]

\multirow{5}{*}{Constrained}
& \rule{0pt}{2.7ex}4-L, 4-Y
& 16 & 6 & 532{,}230 & 530{,}945 & 1{,}063{,}175 \\
& 5-L, 5-T
& 25 & 8 & 535{,}048 & 533{,}249 & 1{,}068{,}297 \\
& 6-L, 6-T, 6-Y
& 36 & 10 & 538{,}378 & 536{,}065 & 1{,}074{,}443 \\
& 7-L, 7-Y, 7-T
& 49 & 12 & 542{,}220 & 539{,}393 & 1{,}081{,}613 \\
& 8-H, 8-F, 8-T1, 8-T2
& 64 & 14 & 546{,}574 & 543{,}233 & 1{,}089{,}807 \\[2pt]
\hline
\end{tabular}

\caption{Inventory of the trained \texttt{AlphaCNOT} models used in the
main experiments. All models use a width-256 ResMLP with three residual
blocks and nine hidden affine layers in total. Parameter counts refer to
one independent policy-value pair.}
\label{tab:alphacnot-model-inventory}
\end{table*}

\section{QAOA as a phase polynomial}\label{Appx:qaoa}
We consider two well-known problems. Given an undirected graph $G=(V,E)$, MaxCut asks for a partition of $V$ into two sets that maximizes the number of edges crossing it, and it is encoded by the cost Hamiltonian
\begin{equation}
    H_{\mathrm{MC}}=\frac{1}{2}\sum_{(i,j)\in E}(I-Z_iZ_j).
    \label{eq:ricca_maxcut}
\end{equation}
Low Autocorrelation Binary Sequences (LABS), on the other hand, asks for a spin sequence $s\in\{-1,+1\}^n$ that minimizes
\begin{equation}
    E_{\mathrm{LABS}}(s)=\sum_{k=1}^{n-1}
    \left(\sum_{i=1}^{n-k}s_i s_{i+k}\right)^2,
    \label{eq:ricca_labs}
\end{equation}
and its cost Hamiltonian $H_{\mathrm{LABS}}$ is obtained by replacing each $s_i$ with $Z_i$~\citep{shaydulin2024labs}. In both cases, $H_C$ is a sum of products of $Z$ operators: two-body ($ZZ$) terms for MaxCut, and both two-body and four-body ($ZZZZ$) terms for LABS, while constant terms only contribute a global phase and are dropped. Therefore, the cost evolution $e^{-\mathrm{i}\gamma H_C}$ of a QAOA layer is a sequence of $Z$-phase gadgets (Eq.~\ref{eq:ricca_phase_gadget}), where each term $c\prod_{j\in S}Z_j$ gives a gadget with angle $2\gamma c$. The mixer $e^{-\mathrm{i}\beta\sum_j X_j}$ simply adds single-qubit $X$-phase gadgets with angle $2\beta$. 

As described in Section~\ref{sub3:phasepolynomials}, ParitySynth pushes the uncomputing \CNOTs to the end of the circuit, and returns the prefix $U_{\mathrm{pre}}$ together with the trailing circuit $U_{\mathrm{tail}}$. Note that the original tail leaves a lot of room for improvement.

Figure~\ref{fig:ricca_tail_example} shows the tail of 14 \CNOTs returned by ParitySynth with parity matrix
\begin{equation*}
    M_{\mathrm{tail}}=
    \begin{pmatrix}
        0 & 0 & 1 & 0\\
        0 & 1 & 1 & 0\\
        1 & 0 & 0 & 0\\
        0 & 0 & 0 & 1
    \end{pmatrix},
\end{equation*}
i.e., the outputs are $(x_3,\,x_2\oplus x_3,\,x_1,\,x_4)$. In the fixed-order setting, RowCol needs $8$ \CNOTs, since $x_1$ and $x_3$ have to be exchanged across the line. If the outputs can be permuted, instead, a single $\CNOT(3,2)$ is enough: the wires carry $(x_1,\,x_2\oplus x_3,\,x_3,\,x_4)$, which is $M_{\mathrm{tail}}x$ up to exchanging the labels of the first and third qubit.

% RICCA: example taken from case n4_L_labs_0 of the baseline dataset
\begin{figure}[htbp]
    \centering

    \begin{quantikz}[row sep=0.25cm, column sep=0.22cm]
        \lstick{$x_1$} & \ctrl{1} &  &  & \targ{} &  &  &  & \ctrl{1} &  &  &  & \targ{} &  &  & \rstick{$x_3$} \\
        \lstick{$x_2$} & \targ{} & \ctrl{1} & \targ{} & \ctrl{-1} & \ctrl{1} &  & \ctrl{1} & \targ{} & \ctrl{1} & \ctrl{1} &  & \ctrl{-1} & \targ{} & \ctrl{1} & \rstick{$x_2\oplus x_3$} \\
        \lstick{$x_3$} &  & \targ{} & \ctrl{-1} &  & \targ{} & \targ{} & \targ{} &  & \targ{} & \targ{} & \targ{} &  & \ctrl{-1} & \targ{} & \rstick{$x_1$} \\
        \lstick{$x_4$} &  &  &  &  &  & \ctrl{-1} &  &  &  &  & \ctrl{-1} &  &  &  & \rstick{$x_4$}
    \end{quantikz}
    \caption{Original tail returned by ParitySynth (14 \CNOTs).}
    \label{fig:ricca_tail_example}
\end{figure}

\section{Topological Structures}\label{Appx:topology}
In this final section, we present the topologies used in the comparative analysis with the RL-CL method \citep{kremer2024}. The nodes of the graphs represent the physical qubits, while the edges denote the connectivity constraints of the specific hardware architecture. In all evaluated cases, the graphs are considered undirected, i.e., any connection $(i, j)$ permits the execution of both $\CNOT_{ij}$ and $\CNOT_{ji}$ operations without distinction.

\begin{figure*}[h]
    \centering
    \small

    \begin{subfigure}[b]{0.22\textwidth}
        \centering
        \includegraphics[width=\textwidth]{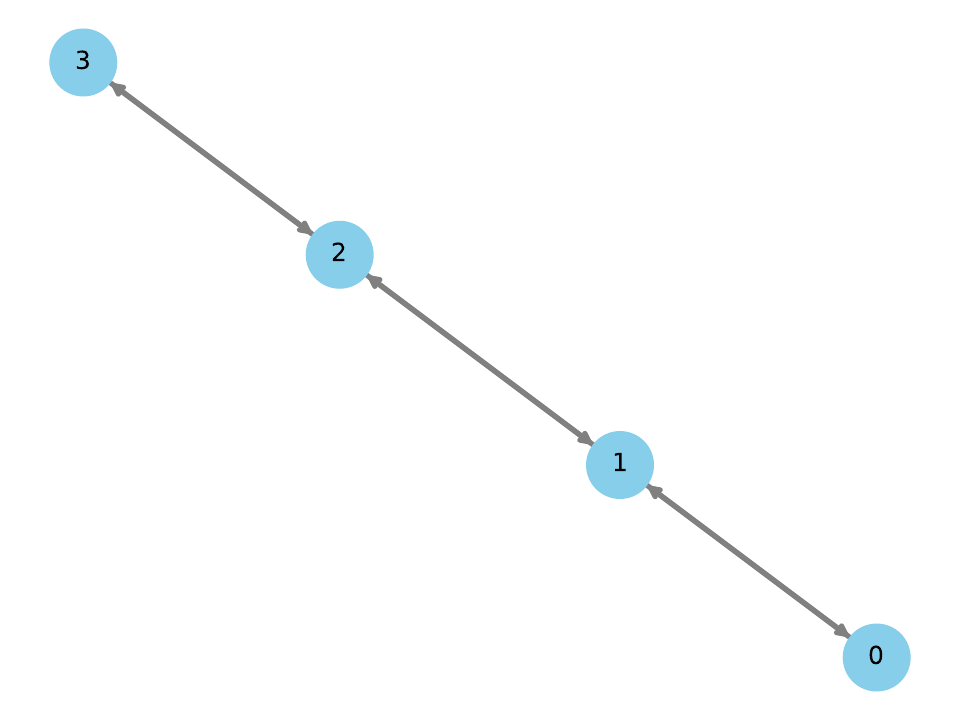}
        \caption{4-L}
    \end{subfigure}\hfill
    \begin{subfigure}[b]{0.22\textwidth}
        \centering
        \includegraphics[width=\textwidth]{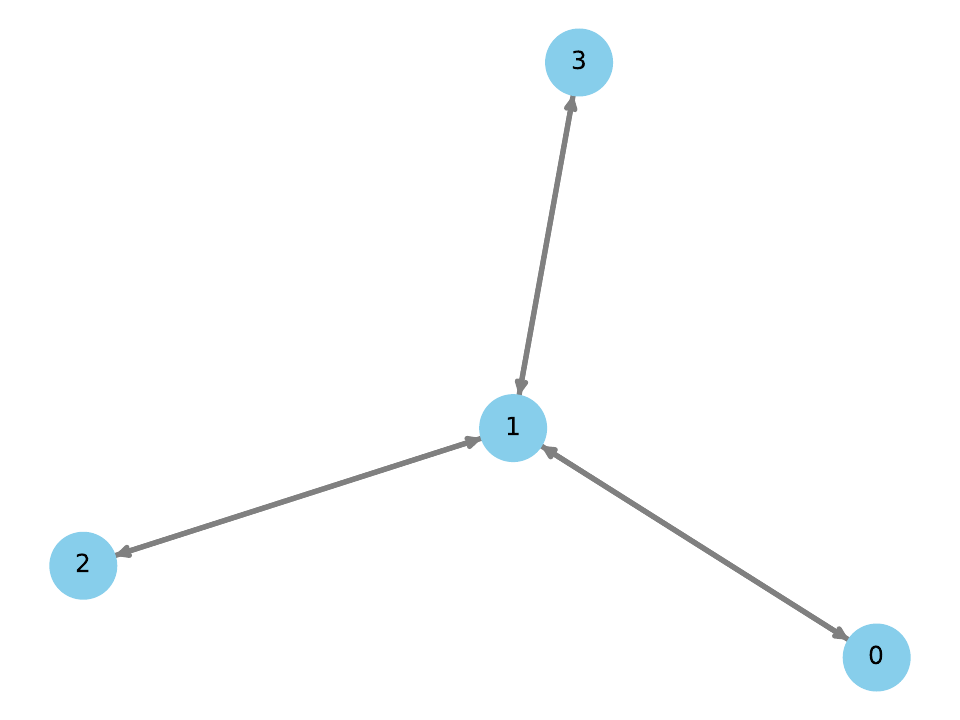}
        \caption{4-Y}
    \end{subfigure}\hfill
    \begin{subfigure}[b]{0.22\textwidth}
        \centering
        \includegraphics[width=\textwidth]{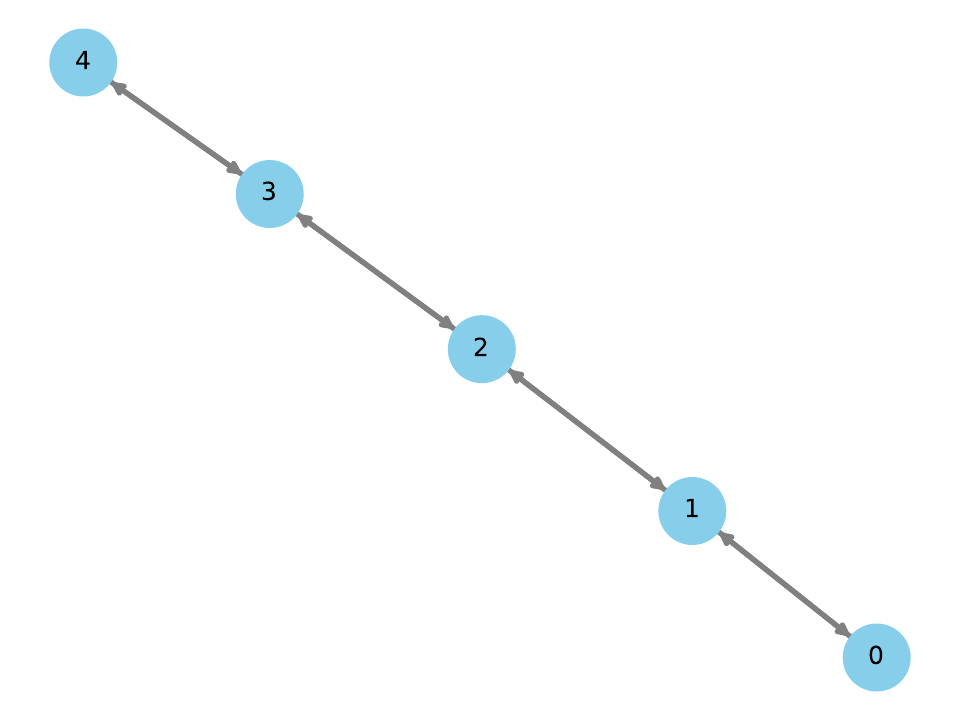}
        \caption{5-L}
    \end{subfigure}\hfill
    \begin{subfigure}[b]{0.22\textwidth}
        \centering
        \includegraphics[width=\textwidth]{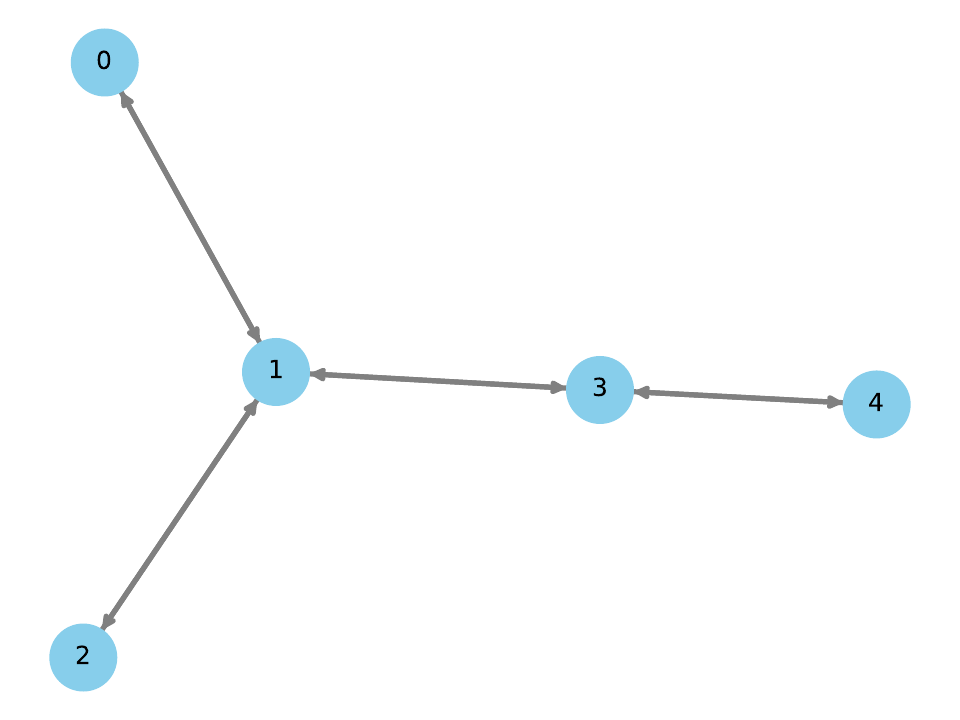}
        \caption{5-T}
    \end{subfigure}

    \vspace{1.5em}

    \begin{subfigure}[b]{0.22\textwidth}
        \centering
        \includegraphics[width=\textwidth]{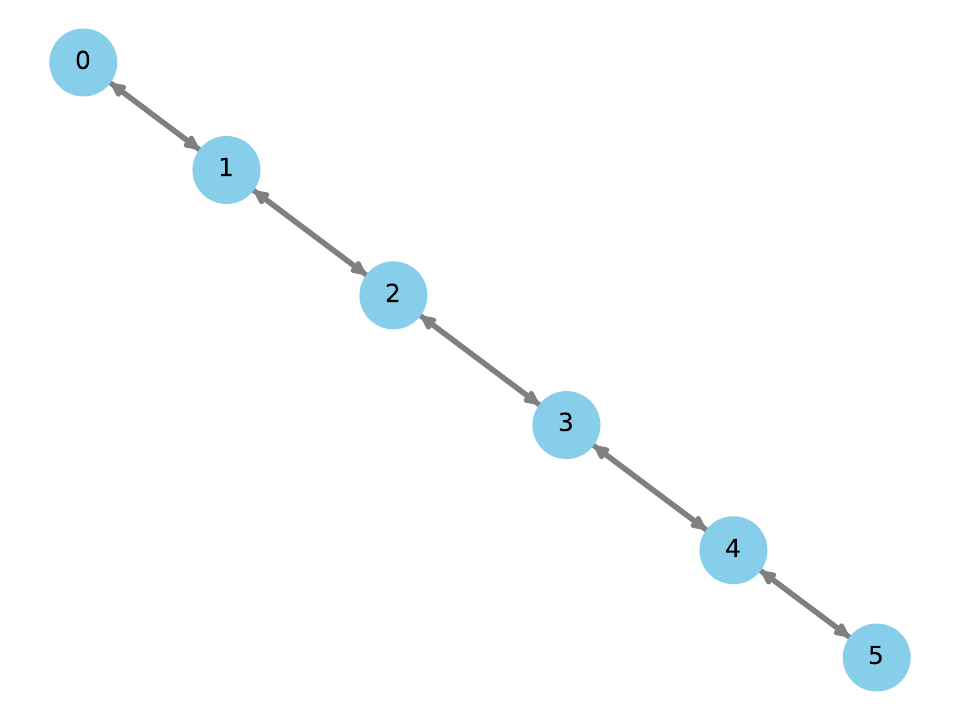}
        \caption{6-L}
    \end{subfigure}\hfill
    \begin{subfigure}[b]{0.22\textwidth}
        \centering
        \includegraphics[width=\textwidth]{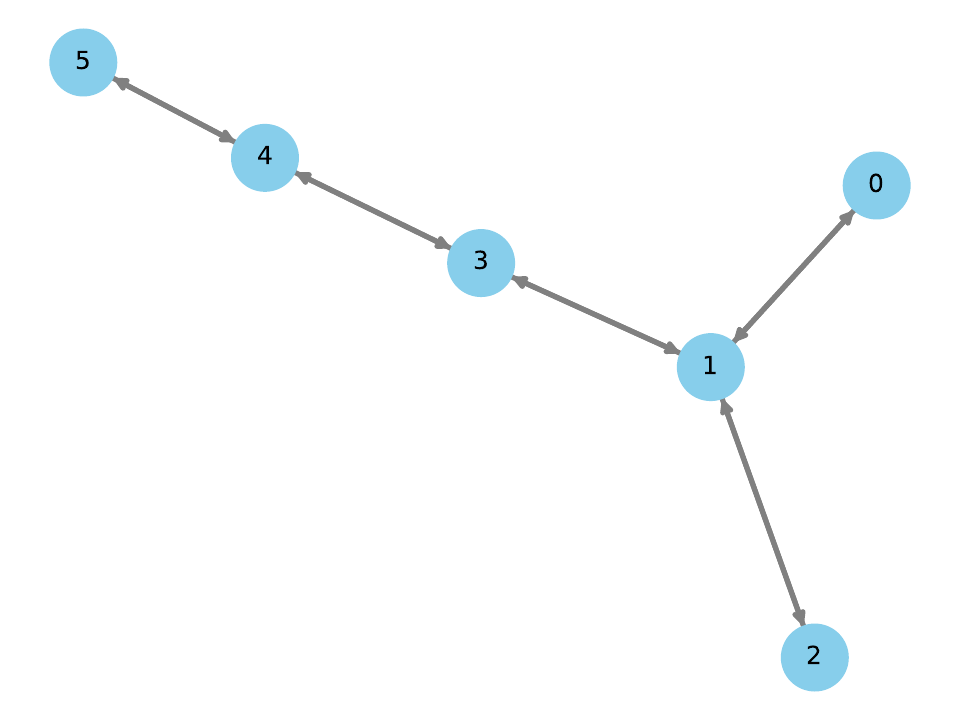}
        \caption{6-T}
    \end{subfigure}\hfill
    \begin{subfigure}[b]{0.22\textwidth}
        \centering
        \includegraphics[width=\textwidth]{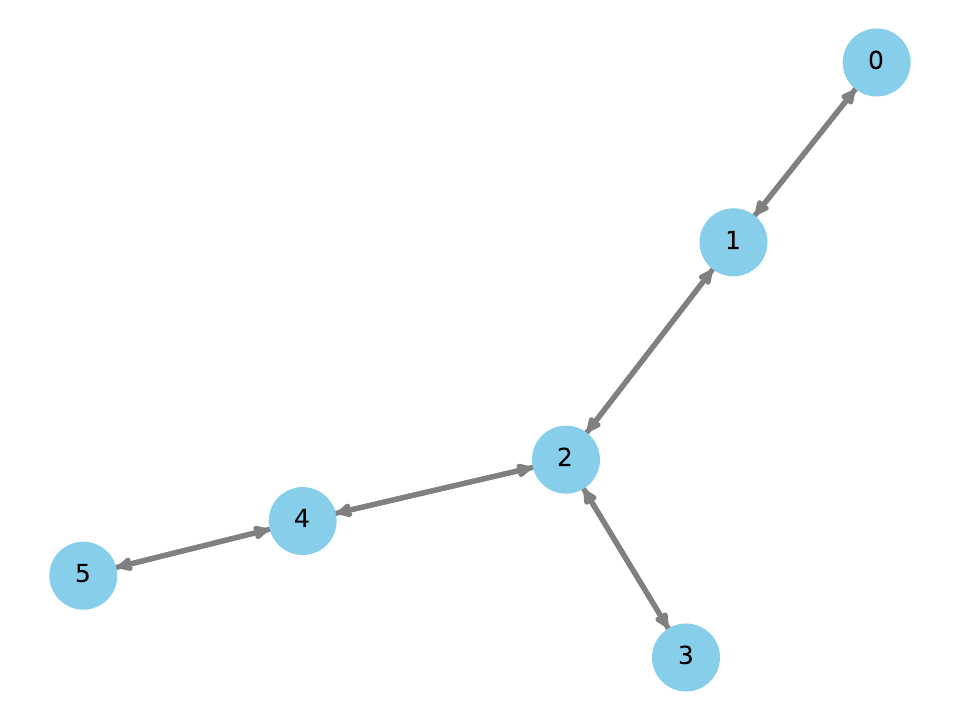}
        \caption{6-Y}
    \end{subfigure}\hfill
    \begin{subfigure}[b]{0.22\textwidth}
        \centering
        \includegraphics[width=\textwidth]{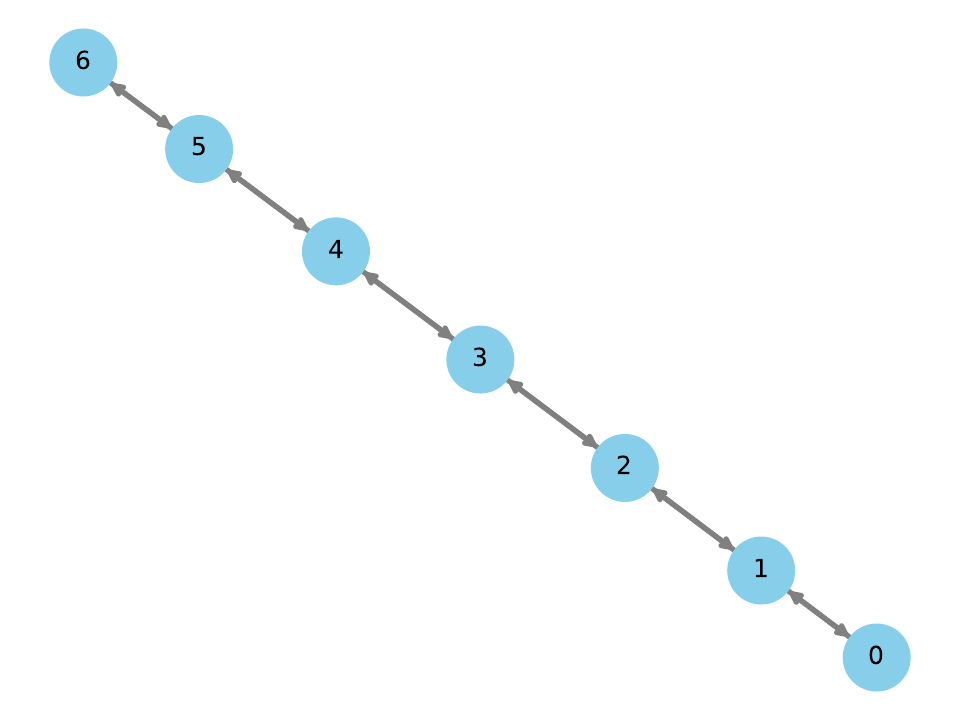}
        \caption{7-L}
    \end{subfigure}\hfill

    \vspace{1.5em}
    \begin{subfigure}[b]{0.22\textwidth}
        \centering
        \includegraphics[width=\textwidth]{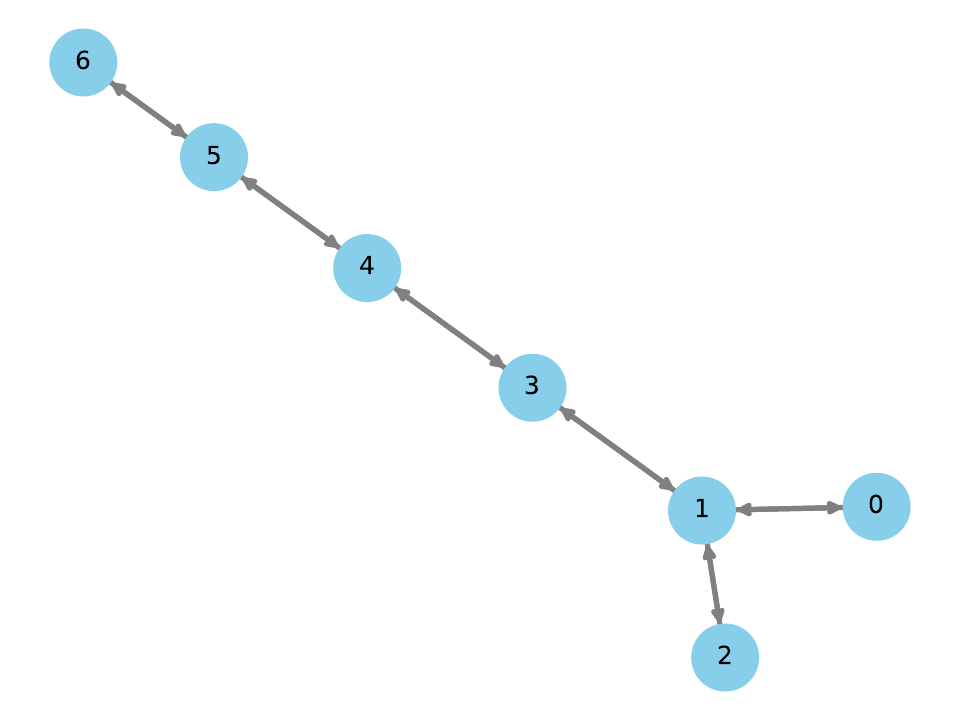}
        \caption{7-T}
    \end{subfigure}
    \begin{subfigure}[b]{0.22\textwidth}
        \centering
        \includegraphics[width=\textwidth]{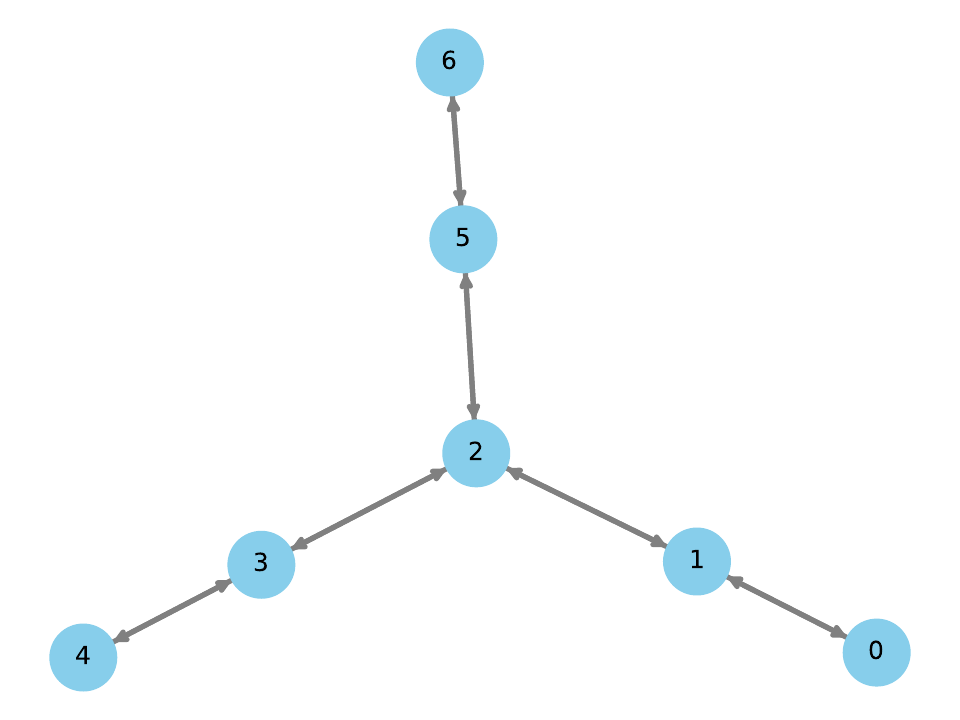}
        \caption{7-Y}
    \end{subfigure}\hfill
    \begin{subfigure}[b]{0.22\textwidth}
        \centering
        \includegraphics[width=\textwidth]{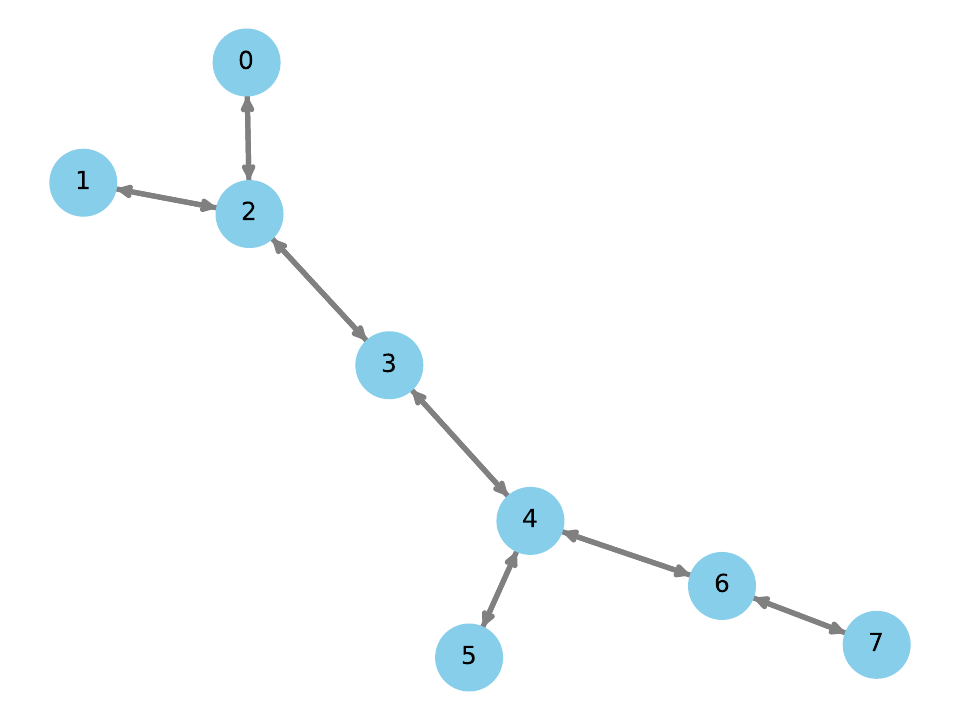}
        \caption{8-H}
    \end{subfigure}\hfill
    \begin{subfigure}[b]{0.22\textwidth}
        \centering
        \includegraphics[width=\textwidth]{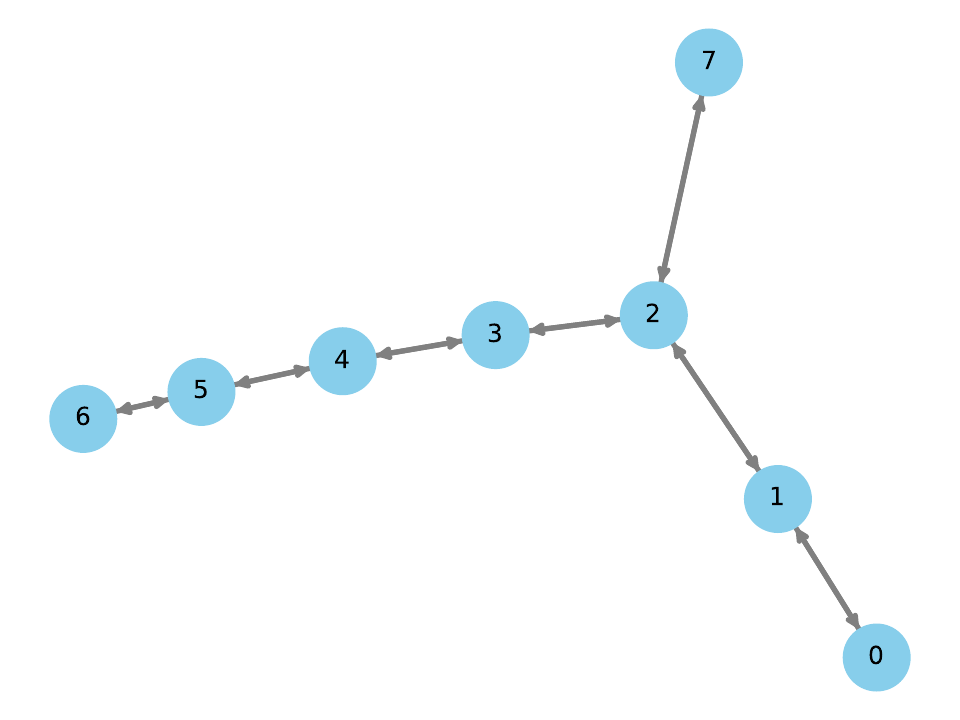}
        \caption{8-F}
    \end{subfigure}\hfill

    \vspace{1.5em}
    \begin{subfigure}[b]{0.22\textwidth}
        \centering
        \includegraphics[width=\textwidth]{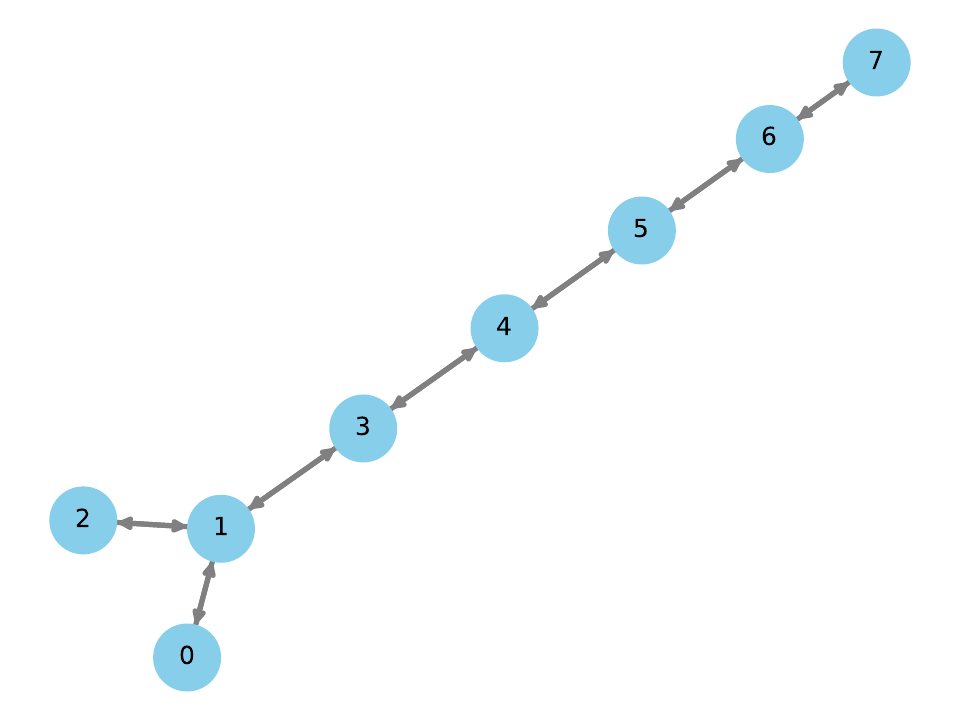}
        \caption{8-T1}
    \end{subfigure}
    \begin{subfigure}[b]{0.22\textwidth}
        \centering
        \includegraphics[width=\textwidth]{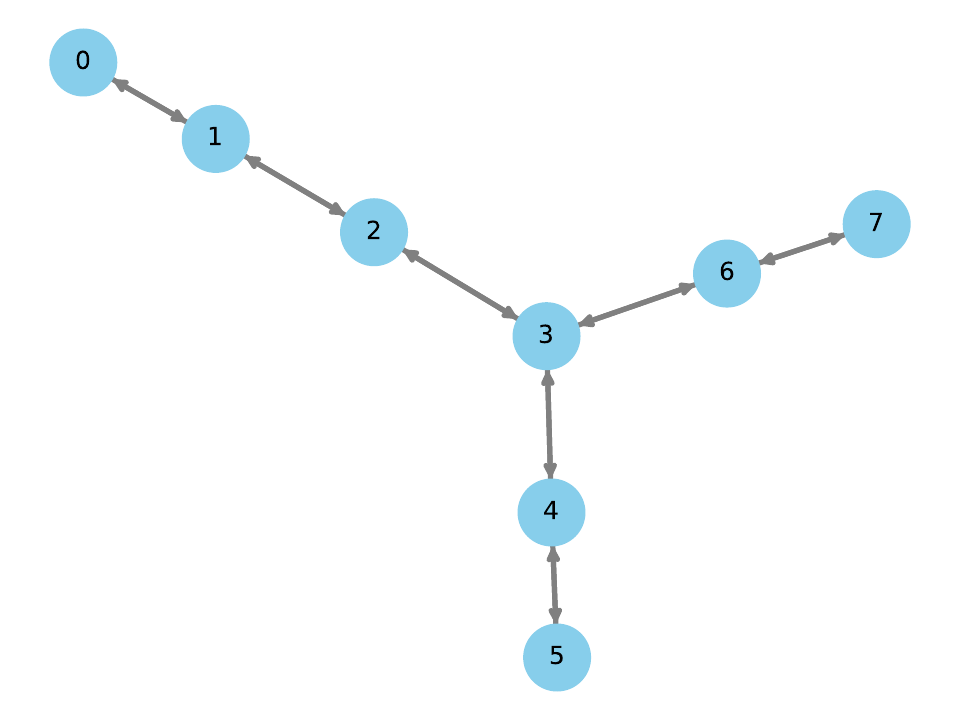}
        \caption{8-T2}
    \end{subfigure}

    \caption{Complete set of the fourteen hardware topologies considered in the study, spanning configurations from four to eight qubits. The nomenclature is based on the shape and is consistent with what is provided in \cite{kremer2024}.}
    \label{fig:appendix_topologies}
\end{figure*}

\end{appendices}

\end{document}